\pdfoutput=1
\PassOptionsToPackage{table}{xcolor}
\documentclass{article}
\makeatletter

\usepackage{eso-pic} %
\RequirePackage{fancyhdr}
\RequirePackage{natbib}

\setcitestyle{authoryear,round,citesep={;},aysep={,},yysep={;}}

\newif\ificlrfinal
\iclrfinalfalse
\def\iclrfinalcopy{\iclrfinaltrue}
\font\iclrtenhv  = phvb at 8pt

\def\addcontentsline#1#2#3{}

\def\maketitle{\par
\begingroup
   \def\thefootnote{\fnsymbol{footnote}}
   \def\@makefnmark{\hbox to 0pt{$^{\@thefnmark}$\hss}} %
   \long\def\@makefntext##1{\parindent 1em\noindent
                            \hbox to1.8em{\hss $\m@th ^{\@thefnmark}$}##1}
   \@maketitle \@thanks
\endgroup
\setcounter{footnote}{0}
\let\maketitle\relax \let\@maketitle\relax
\gdef\@thanks{}\gdef\@author{}\gdef\@title{}\let\thanks\relax}

\usepackage{fancyhdr}
\def\@maketitle{\vbox{\hsize\textwidth
{\LARGE\sc \@title\par}
\ificlrfinal
    \lhead{}
    \def\And{\end{tabular}\hfil\linebreak[0]\hfil
            \begin{tabular}[t]{l}\bf\rule{\z@}{24pt}\ignorespaces}%
  \def\AND{\end{tabular}\hfil\linebreak[4]\hfil
            \begin{tabular}[t]{l}\bf\rule{\z@}{24pt}\ignorespaces}%
    \begin{tabular}[t]{l}\bf\rule{\z@}{24pt}\@author\end{tabular}%
\else
       \lhead{Under review as a conference paper at ICLR 2027}
   \def\And{\end{tabular}\hfil\linebreak[0]\hfil
            \begin{tabular}[t]{l}\bf\rule{\z@}{24pt}\ignorespaces}%
  \def\AND{\end{tabular}\hfil\linebreak[4]\hfil
            \begin{tabular}[t]{l}\bf\rule{\z@}{24pt}\ignorespaces}%
    \begin{tabular}[t]{l}\bf\rule{\z@}{24pt}Anonymous authors\\Paper under double-blind review\end{tabular}%
\fi
\vskip 0.3in minus 0.1in}}

\renewenvironment{abstract}{\vskip.075in\centerline{\large\sc
Abstract}\vspace{0.5ex}\begin{quote}}{\par\end{quote}\vskip 1ex}

\def\section{\@startsection {section}{1}{\z@}{-2.0ex plus
    -0.5ex minus -.2ex}{1.5ex plus 0.3ex
minus0.2ex}{\large\sc\raggedright}}

\def\subsection{\@startsection{subsection}{2}{\z@}{-1.8ex plus
-0.5ex minus -.2ex}{0.8ex plus .2ex}{\normalsize\sc\raggedright}}
\def\subsubsection{\@startsection{subsubsection}{3}{\z@}{-1.5ex
plus      -0.5ex minus -.2ex}{0.5ex plus
.2ex}{\normalsize\sc\raggedright}}
\def\paragraph{\@startsection{paragraph}{4}{\z@}{1.5ex plus
0.5ex minus .2ex}{-1em}{\normalsize\bf}}
\def\subparagraph{\@startsection{subparagraph}{5}{\z@}{1.5ex plus
  0.5ex minus .2ex}{-1em}{\normalsize\sc}}

\skip\footins 9pt plus 4pt minus 2pt
\def\footnoterule{\kern-3pt \hrule width 12pc \kern 2.6pt }
\leftmargini\leftmargin \leftmarginii 2em
\def\@listi{\leftmargin\leftmargini}
\def\@listii{\leftmargin\leftmarginii
   \labelwidth\leftmarginii\advance\labelwidth-\labelsep
   \topsep 2pt plus 1pt minus 0.5pt
   \parsep 1pt plus 0.5pt minus 0.5pt
   \itemsep \parsep}
\def\@listiii{\leftmargin\leftmarginiii
    \labelwidth\leftmarginiii\advance\labelwidth-\labelsep
    \topsep 1pt plus 0.5pt minus 0.5pt
    \parsep \z@ \partopsep 0.5pt plus 0pt minus 0.5pt
    \itemsep \topsep}
\def\@listiv{\leftmargin\leftmarginiv
     \labelwidth\leftmarginiv\advance\labelwidth-\labelsep}
\def\@listv{\leftmargin\leftmarginv
     \labelwidth\leftmarginv\advance\labelwidth-\labelsep}
\def\@listvi{\leftmargin\leftmarginvi
     \labelwidth\leftmarginvi\advance\labelwidth-\labelsep}

\belowdisplayskip \abovedisplayskip
\def\normalsize{\@setsize\normalsize{11pt}\xpt\@xpt}
\def\small{\@setsize\small{10pt}\ixpt\@ixpt}
\def\footnotesize{\@setsize\footnotesize{10pt}\ixpt\@ixpt}
\def\scriptsize{\@setsize\scriptsize{8pt}\viipt\@viipt}
\def\tiny{\@setsize\tiny{7pt}\vipt\@vipt}
\def\large{\@setsize\large{14pt}\xiipt\@xiipt}
\def\Large{\@setsize\Large{16pt}\xivpt\@xivpt}
\def\LARGE{\@setsize\LARGE{20pt}\xviipt\@xviipt}
\def\huge{\@setsize\huge{23pt}\xxpt\@xxpt}
\def\Huge{\@setsize\Huge{28pt}\xxvpt\@xxvpt}

\makeatletter
\newbox\iclrrulerbox
\newcount\iclrrulercount
\newdimen\iclrruleroffset
\newdimen\cv@lineheight
\newdimen\cv@boxheight
\newbox\cv@tmpbox
\newcount\cv@refno
\newcount\cv@tot
\newcount\cv@tmpc@ \newcount\cv@tmpc
\def\fillzeros[#1]#2{\cv@tmpc@=#2\relax\ifnum\cv@tmpc@<0\cv@tmpc@=-\cv@tmpc@\fi
\cv@tmpc=1 %
\loop\ifnum\cv@tmpc@<10 \else \divide\cv@tmpc@ by 10 \advance\cv@tmpc by 1 \fi
   \ifnum\cv@tmpc@=10\relax\cv@tmpc@=11\relax\fi \ifnum\cv@tmpc@>10 \repeat
\ifnum#2<0\advance\cv@tmpc1\relax-\fi
\loop\ifnum\cv@tmpc<#1\relax0\advance\cv@tmpc1\relax\fi \ifnum\cv@tmpc<#1 \repeat
\cv@tmpc@=#2\relax\ifnum\cv@tmpc@<0\cv@tmpc@=-\cv@tmpc@\fi \relax\the\cv@tmpc@}%
\def\makevruler[#1][#2][#3][#4][#5]{\begingroup\offinterlineskip
\textheight=#5\vbadness=10000\vfuzz=120ex\overfullrule=0pt%
\global\setbox\iclrrulerbox=\vbox to \textheight{%
{\parskip=0pt\hfuzz=150em\cv@boxheight=\textheight
\cv@lineheight=#1\global\iclrrulercount=#2%
\cv@tot\cv@boxheight\divide\cv@tot\cv@lineheight\advance\cv@tot2%
\cv@refno1\vskip-\cv@lineheight\vskip1ex%
\loop\setbox\cv@tmpbox=\hbox to0cm{{\iclrtenhv\hfil\fillzeros[#4]\iclrrulercount}}%
\ht\cv@tmpbox\cv@lineheight\dp\cv@tmpbox0pt\box\cv@tmpbox\break
\advance\cv@refno1\global\advance\iclrrulercount#3\relax
\ifnum\cv@refno<\cv@tot\repeat}}\endgroup}%
\makeatother

\def\iclrruler#1{\makevruler[12pt][#1][1][3][0.993\textheight]\usebox{\iclrrulerbox}}
\AddToShipoutPicture{%
\ificlrfinal\else
\iclrruleroffset=\textheight
\advance\iclrruleroffset by -3.7pt
  \color[rgb]{.7,.7,.7}
  \AtTextUpperLeft{%
    \put(\LenToUnit{-35pt},\LenToUnit{-\iclrruleroffset}){%
      \iclrruler{\iclrrulercount}}
  }
\fi
}

\makeatother
\usepackage{times}
\usepackage{amsmath,amssymb,amsthm,mathtools}
\usepackage{graphicx,booktabs,tabularx,array,multirow}
\usepackage{algorithm,algpseudocode}
\usepackage{xcolor}
\usepackage{colortbl}
\usepackage{enumitem}
\usepackage{placeins}
\usepackage{hyperref}
\usepackage{url}
\usepackage{xspace}
\hypersetup{colorlinks=true,linkcolor=blue!55!black,citecolor=blue!55!black,urlcolor=blue!55!black}

\title{DiffusionRVGE: Reinforcement Learning via\\Regression-Based Value Gradient Estimation}
\author{Jiayuan Wang \\
Qiuzhen College, Tsinghua University \\
Beijing 100084, China \\
\texttt{jy-wang25@mails.tsinghua.edu.cn}
\And
Tongyao Pang\thanks{Corresponding author.} \\
Yau Mathematical Sciences Center \\
Shuangqing Complex, Tsinghua University \\
Beijing 100084, China \\
\texttt{typang@tsinghua.edu.cn}
}
\iclrfinalcopy
\newcommand{\method}{DiffusionRVGE\xspace}
\newcommand{\E}{\mathbb{E}}
\newcommand{\R}{\mathbb{R}}
\newcommand{\Cov}{\operatorname{Cov}}
\newcommand{\sg}{\operatorname{sg}}
\newcommand{\row}{\operatorname{row}}

\newcommand{\dd}{\mathrm{d}}
\newcommand{\vect}[1]{\mathbf{#1}}
\newcommand{\vecg}[1]{\boldsymbol{#1}}
\newcommand{\btheta}{\boldsymbol{\theta}}
\newcommand{\bthetaold}{\boldsymbol{\theta}_{\mathrm{old}}}
\newtheorem{theorem}{Theorem}
\newtheorem{proposition}[theorem]{Proposition}

\theoremstyle{remark}

\begin{document}
\maketitle
\begin{abstract}
Reinforcement learning improves diffusion models using rewards on generated images, but repeated denoising and gradient estimation make training expensive. We introduce \method, an efficient method that constructs useful update directions by regressing centered rewards on centered final denoised images. It requires neither reward derivatives nor adjoint backpropagation through the sampling trajectory. A unified first-variation formulation identifies the value-gradient target with or without a trajectory-KL penalty. A population analysis under Gaussian noising explains the regression geometry, while a subspace analysis motivates clean-image coordinates. The implementation adapts rollout bandwidth and query weights to reward continuity: local rollouts suit smooth rewards, whereas broader groups provide informative contrasts for discrete rewards. Trained jointly on five rewards, one SD3.5-Medium model matches or exceeds DiffusionNFT's joint model on seven of eight metrics, reaching GenEval $0.94$, OCR $0.92$, and PickScore $23.82$ at about one fifth of its estimated training compute. In single-reward training, \method reaches matched reward thresholds $2.1$--$6.7\times$ faster than reproduced DiffusionNFT on identical hardware. Ablations support the clean-image estimator, direct direction transfer, and reward-dependent rollout choices.
Code is available at \url{https://github.com/jy-wang-tsinghua/DIFFUSIONRVGE}.
\end{abstract}

\newcommand{\oPickGE}{0.65}
\newcommand{\oPickOCR}{0.63}
\newcommand{\oPickPick}{23.99}
\newcommand{\oPickClip}{0.275}
\newcommand{\oPickHPS}{0.328}
\newcommand{\oPickAes}{6.28}
\newcommand{\oPickImg}{1.38}
\newcommand{\oPickUni}{3.45}
\newcommand{\oGEGE}{0.98}
\newcommand{\oGEOCR}{0.48}
\newcommand{\oGEPick}{21.94}
\newcommand{\oGEClip}{0.287}
\newcommand{\oGEHPS}{0.252}
\newcommand{\oGEAes}{5.11}
\newcommand{\oGEImg}{0.77}
\newcommand{\oGEUni}{3.07}

\newcommand{\oOCRGE}{0.15}
\newcommand{\oOCROCR}{0.98}
\newcommand{\oOCRPick}{20.32}
\newcommand{\oOCRClip}{0.236}
\newcommand{\oOCRHPS}{0.198}
\newcommand{\oOCRAes}{4.65}
\newcommand{\oOCRImg}{-0.93}
\newcommand{\oOCRUni}{2.00}

\newcommand{\oOCRfGE}{0.51}
\newcommand{\oOCRfOCR}{0.97}
\newcommand{\oOCRfPick}{22.17}
\newcommand{\oOCRfClip}{0.276}
\newcommand{\oOCRfHPS}{0.292}
\newcommand{\oOCRfAes}{5.64}
\newcommand{\oOCRfImg}{0.87}
\newcommand{\oOCRfUni}{3.02}
\newcommand{\oOCRfLayout}{0.51}

\newcommand{\oMultiGE}{0.94}
\newcommand{\oMultiOCR}{0.92}
\newcommand{\oMultiPick}{23.82}
\newcommand{\oMultiClip}{0.303}
\newcommand{\oMultiHPS}{0.339}
\newcommand{\oMultiAes}{5.95}
\newcommand{\oMultiImg}{1.55}
\newcommand{\oMultiUni}{3.50}
\newcommand{\oMultiGPUh}{404}

\newcommand{\nGEGE}{0.90}     %
\newcommand{\nGEOCR}{0.42}
\newcommand{\nGEPick}{22.03}
\newcommand{\nGEClip}{0.276}
\newcommand{\nGEHPS}{0.256}
\newcommand{\nGEAes}{5.20}
\newcommand{\nGEImg}{0.62}
\newcommand{\nGEUni}{2.99}
\newcommand{\nOCRGE}{0.25}
\newcommand{\nOCROCR}{0.96}   %
\newcommand{\nOCRPick}{20.94}
\newcommand{\nOCRClip}{0.254}
\newcommand{\nOCRHPS}{0.206}
\newcommand{\nOCRAes}{4.94}
\newcommand{\nOCRImg}{-0.53}
\newcommand{\nOCRUni}{2.41}
\newcommand{\nPickGE}{0.63}
\newcommand{\nPickOCR}{0.56}
\newcommand{\nPickPick}{23.50}  %
\newcommand{\nPickClip}{0.266}
\newcommand{\nPickHPS}{0.299}
\newcommand{\nPickAes}{6.07}
\newcommand{\nPickImg}{1.19}
\newcommand{\nPickUni}{3.24}

\section{Introduction}
\label{sec:intro}
Diffusion and flow matching models generate images by transforming noise into data \citep{ho2020ddpm,song2021score,lipman2023flow,esser2024sd3}. Reinforcement learning (RL) adapts these generators to scalar rewards measuring preferences, text accuracy, or compositional correctness. Methods including DDPO, Flow-GRPO, and DiffusionNFT improve generation through reward-based fine-tuning \citep{black2023ddpo,liu2025flowgrpo,zheng2026nft}. Training remains expensive: each update requires image generation and an estimate of how model changes affect the final reward. Our aim is to obtain effective update directions at substantially lower training cost.

We introduce \method, which constructs reward-informed update directions by regression on final denoised images. To motivate the estimator, consider the Gaussian noising path
\begin{equation}
 \vect{X}_t=\alpha_t\vect{Z}+\sigma_t\vecg{\epsilon},
 \qquad \vect{Z}\sim\rho,\quad
 \vecg{\epsilon}\sim\mathcal N(0,I),
 \label{eq:intro_noising}
\end{equation}
where $\vect{Z}$ and $\vecg{\epsilon}$ are independent, $\vect{Z}$ is a clean image or latent, and generation reverses the noising process. Let $P_{\btheta}$ denote the law of a stochastic denoising trajectory. A common RL objective, $\E_{P_{\btheta}}R(\vect{Z})-\tau\operatorname{KL}(P_{\btheta}\Vert P_{\mathrm{ref}})$, balances reward against deviation from a reference model. For $\tau>0$, under an independent-noise initialization, its optimal-control formulation identifies a reward-tilted terminal target $\pi^*\propto\rho e^{R/\tau}$. The policy first variation gives a common cross-term plus quadratic-penalty loss for every $\tau\geq0$, including unregularized reward optimization. This supplies a target for efficient approximation, rather than requiring differentiation through every denoising step.

Our estimator fits centered reward differences to clean-image contrasts using normalized ridge regression. Under exact conditioning in the Gaussian noising model, the population normal equations connect clean-image regression to a covariance-preconditioned value gradient. A subspace analysis characterizes how denoised coordinates improve directional estimation, and our alignment experiments demonstrate this advantage in the evaluated models. We transfer the fitted directions directly to model updates without an adjoint backward pass.

Regression uses reward values and benefits from informative variation across the rollout bank. Smooth scores permit local perturbations; discrete scores can remain constant across these perturbations and benefit from independent-seed groups. We therefore adapt rollout bandwidth and query weights to reward continuity while retaining the same centered squared-regression estimator. This gives one practical method for both continuous preferences and discrete task scores.

Our contributions are:
\begin{itemize}[leftmargin=*,itemsep=1pt,topsep=2pt]
\item \textbf{Efficient regression-based updates.} Clean-image regression constructs update directions without reward derivatives or adjoint backpropagation, guided by gradient matching and supported by directional analysis and empirical validation.
\item \textbf{Reward-adaptive implementation.} Local trajectory banks and broader rollout groups share a centered regression formulation, with reward-dependent query weights validated by ablations.
\item \textbf{Strong results at lower compute.} Our joint model matches or exceeds DiffusionNFT on seven of eight metrics at about one fifth of its estimated training compute. Single-reward runs reach matched thresholds $2.1$--$6.7\times$ faster on identical hardware.
\end{itemize}

\section{Related work}
\label{sec:related}
\paragraph{RL for diffusion and flow models.}
DDPO and Flow-GRPO optimize stochastic denoising policies \citep{black2023ddpo,liu2025flowgrpo}; DiffusionNFT uses positive and negative generations in forward-process matching \citep{zheng2026nft}. DPOK, Diffusion-DPO, DanceGRPO, and Centered Reward Distillation provide other reward- or preference-based objectives \citep{fan2023dpok,wallace2023diffusiondpo,xue2025dance,zhu2026crd}. A path-space perspective identifies value-gradient estimation as a shared component \citep{xu2026designing}. DRaFT differentiates rewards through sampling, whereas FDFO transfers paired reward differences into velocity updates \citep{clark2023draft,mcallister2026fdfo}. Our joint regression accounts for correlations among clean-image differences. Temporal and stepwise methods change branching or credit assignment \citep{he2025tempflow,savani2026stepwise,zhou2025granular,chen2025superflow}; our bandwidth choices address reward continuity and discreteness.

\paragraph{Regression and controlled dynamics.}
Adjoint Matching learns stochastic-control targets through adjoint regression \citep{domingo2024adjoint}. Reward Score Matching and path-space analyses relate diffusion RL to reward-conditioned score or value gradients \citep{lee2026rsm,xu2026designing}. Our analysis distinguishes ordinary reward regression from the soft-value oracle of trajectory-KL control. Dimension-dependent derivative-free estimation and subspace-guided search \citep{nesterov2017random,maheswaranathan2019guided} motivate the clean-coordinate comparison.

\section{Gradient matching with clean-image regression}
\label{sec:theory}
We first identify the RL update target, then study the geometry of clean-image regression. Population and subspace analyses explain the role of preconditioning and the choice of coordinates; the corresponding proofs and technical conditions are in Appendices~\ref{app:pathkl_control} and~\ref{app:preconditioning}.

\subsection{The gradient-matching objective}
\label{sec:control}
Use generation time $s=T-t$ and a reference diffusion with drift $\vect{b}_{\mathrm{ref}}$ and noise coefficient $\gamma_s>0$. A learned correction $\vect{u}_{\btheta}$ defines
\begin{equation}
 \dd\vect{Y}_s=(\vect{b}_{\mathrm{ref}}+\vect{u}_{\btheta})(\vect{Y}_s,s)\dd s+\gamma_s\dd\vect{B}_s,
 \qquad J_\tau=\E_{P_{\btheta}}R(\vect{Y}_T)-\tau\operatorname{KL}(P_{\btheta}\Vert P_{\mathrm{ref}}).
 \label{eq:sde_objective}
\end{equation}
For $\tau>0$, tilting reference paths by $e^{R/\tau}$ gives a law $Q^*$ with terminal distribution $\pi^*\propto e^{R/\tau}\rho$ and
\begin{equation}
 J_\tau=\tau\log Z_\tau-\tau\operatorname{KL}(P_{\btheta}\Vert Q^*),
 \qquad Z_\tau=\E_\rho e^{R/\tau}.
 \label{eq:intro_control}
\end{equation}
Under the independent-noise initialization, this identifies the desired terminal target. Define the soft value $W_t=\tau\log H_t$, where $H_t(\vect{x})=\E_{\mathrm{ref}}[e^{R(\vect{Z})/\tau}\mid\vect{X}_t=\vect{x}]$.

\begin{theorem}[Gradient matching for trajectory-KL RL]
\label{thm:pathkl}
For $\tau>0$, reference and controlled processes with the same initial law and diffusion coefficient satisfy, under the regularity conditions in Appendix~\ref{app:pathkl_control},
\begin{equation}
 C_\tau-J_\tau=\frac{\tau}{2}\E_{P_{\btheta}}\int_0^T\frac{\|\vect{u}_{\btheta}-\vect{u}_s^*\|^2}{\gamma_s^2}\dd s,
 \qquad \vect{u}_s^*=\frac{\gamma_s^2}{\tau}\nabla W_{T-s},
 \label{eq:control_matching}
\end{equation}
where $C_\tau=\E[W_T(\vect{Y}_0)]$ is independent of $\btheta$.
\end{theorem}
For a common local update valid for every $\tau\geq0$, use the continuation value of the frozen collection policy $\bthetaold$:
\begin{equation}
 F_t^{\bthetaold,\tau}(\vect{x})=
 \E_{P_{\bthetaold}}\!\left[R(\vect{Y}_T)
 -\frac\tau2\int_{T-t}^T\frac{\|\vect{u}_{\bthetaold}(\vect{Y}_s,s)\|^2}{\gamma_s^2}\dd s
 \,\middle|\,\vect{Y}_{T-t}=\vect{x}\right].
 \label{eq:alltau_value}
\end{equation}
Equivalently, $F_t^{\bthetaold,\tau}=V_t^{\bthetaold}-\tau C_t^{\bthetaold}$, where $V_t^{\bthetaold}$ is expected terminal reward and $C_t^{\bthetaold}$ is expected future control energy. Thus $F_t^{\bthetaold,0}=V_t^{\bthetaold}$; at the optimal policy for $\tau>0$, $F_t^{*,\tau}=W_t$. The same value definition covers both cases.

For the velocity parameterization $\vect{b}_{\btheta}=A_t\vect{x}-2\vect{v}_{\btheta}$ with $A_t=\dot\alpha_t/\alpha_t$, write $\delta\vect{v}_{\btheta}=\vect{v}_{\btheta}-\vect{v}_{\mathrm{ref}}$ and $\nu_t^2=\gamma_{T-t}^2$. Then $\vect{u}_{\btheta}=-2\delta\vect{v}_{\btheta}$. The first variation of $J_\tau$ gives the detached loss
\begin{equation}
 \mathcal L_{\mathrm{GM},\tau}
 =\E_{\mu_{\bthetaold}}\!\left[w_t\left(
 \underbrace{2\langle\delta\vect{v}_{\btheta},\sg[\nabla F_t^{\bthetaold,\tau}]\rangle}_{\text{value-gradient update}}
 +\underbrace{\frac{2\tau}{\nu_t^2}\|\delta\vect{v}_{\btheta}\|^2}_{\text{trajectory-KL penalty}}\right)\right],\quad\tau\geq0.
 \label{eq:gradient_matching_loss}
\end{equation}
With time-sampling compensation $w_t$, its gradient at $\btheta=\bthetaold$ equals $-\nabla_{\btheta}J_\tau|_{\bthetaold}$ (Appendix~\ref{app:alltau}). Here $\sg$ stops gradients. The quadratic term is simply omitted at $\tau=0$; the remaining cross term follows directly from the first variation. This is a local gradient identity, whereas Theorem~\ref{thm:pathkl} characterizes the positive-penalty optimum. Local regression on query-state perturbations estimates $\nabla F_t^{\bthetaold,\tau}$ from sampled rewards minus future control costs (Appendix~\ref{app:alltau}); the latter are absent when $\tau=0$.

\subsection{Population interpretation and directional analysis}
\label{sec:preconditioning}
We study population regression under the exact conditional law induced by the Gaussian noising model in Eq.~\eqref{eq:intro_noising}. Let $V_t(\vect{x})=\E[R(\vect{Z})\mid\vect{X}_t=\vect{x}]$ be its ordinary value. At the reference policy, $\vect{u}_{\bthetaold}=0$ and $F_t^{\bthetaold,\tau}=V_t$ for every $\tau\geq0$. Write $\vect{m}_t=\E[\vect{Z}\mid\vect{x}]$, $\Sigma_t=\Cov(\vect{Z}\mid\vect{x})$, $\vect{c}_t=\Cov(\vect{Z},R\mid\vect{x})$, and $\kappa_t=\alpha_t/\sigma_t^2$. We fit centered rewards to centered clean images:
\begin{equation}
 \vect{g}_t=\kappa_t\operatorname*{arg\,min}_{\vect{u}}\E\!\left[
 \big(R(\vect{Z})-V_t-\vect{u}^{\top}(\vect{Z}-\vect{m}_t)\big)^2\mid\vect{X}_t=\vect{x}\right],
 \label{eq:population_regression}
\end{equation}
choosing the minimum-norm coefficient.
\begin{proposition}[Population geometry of clean-image regression]
\label{thm:preconditioning}
At $\alpha_t,\sigma_t>0$, for Gaussian corruption with finite conditional moments and justified differentiation under the conditional integral,
\begin{equation}
 \underbrace{\nabla_{\vect{x}}V_t=\kappa_t\vect{c}_t}_{\text{value gradient}},
 \qquad
 \underbrace{\vect{g}_t=\kappa_t\Sigma_t^\dagger\vect{c}_t=\Sigma_t^\dagger\nabla_{\vect{x}}V_t}_{\text{preconditioned direction}}.
 \label{eq:preconditioning}
\end{equation}
No derivative of $R$ is needed. The population direction has positive inner product with $\nabla V_t$ whenever the latter is nonzero.
\end{proposition}
For an affine reward with slope $\vect{w}$ on the identifiable clean subspace, $\vect{c}_t=\Sigma_t\vect{w}$ while the population regression coefficient $\vect{g}_t/\kappa_t$ recovers $\vect{w}$. Thus regression accounts for covariance scaling of affine reward slopes on that subspace. For $\tau>0$, the optimal soft value has the analogous covariance formula with $e^{R/\tau}$; its ordinary-to-soft-value residual is discussed in Appendix~\ref{app:control_proof}.

\paragraph{Directional benefit of clean-image coordinates.}
In an isotropic Gaussian clean-subspace model of dimension $k<d$ with an affine reward, consider the same $m\geq2$ Gaussian perturbations and mean-centered observations. With $n=m-1$ independent contrasts, the unregularized minimum-norm fits satisfy
\begin{equation}
 \E\cos^2(\widehat{\vect{h}}_{\mathrm{noisy}},\nabla V_t)=\min(n/d,1),\qquad
 \E\cos^2(\widehat{\vect{g}}_{\mathrm{clean}},\nabla V_t)=\min(n/k,1).
 \label{eq:posterior_dimension}
\end{equation}
In this model, clean-image regression identifies the useful direction from fewer observations. The subspace comparison, proved in Appendix~\ref{app:posterior_dimension}, motivates the choice of clean-image coordinates; Section~\ref{sec:endpoint_angle} evaluates their directional alignment in the trained model. Ridge regularization additionally bounds coefficient sensitivity even for a singular design (Appendix~\ref{app:ridge_stability}). Directly using the fitted clean-image direction avoids an adjoint backward pass; Section~\ref{sec:transport_ablation} compares it with adjoint transport.

A supplementary idealized analysis of Langevin-corrected sampling under additional score-error and mixing assumptions is provided in Appendix~\ref{app:control_proof}.

\section{Reward-adaptive rollout and regression}
\label{sec:method}
\label{sec:practical}
\method constructs update directions by regressing reward differences on clean-image contrasts. The rollout bank determines which contrasts enter the fit, and centered ridge regression combines them into a direction for each query. We adapt rollout bandwidth to the reward's variation and use reward-dependent query weights to determine where the fitted directions are applied.

\paragraph{Reward-dependent rollout bandwidth.}
For smooth rewards, small-noise trajectory banks capture local reward variation by reusing denoised completions along a stochastic rollout. For discrete or nearly piecewise-constant rewards, independent-seed groups provide broader contrasts across reward plateaus. Both constructions produce a bank $\mathcal B_i$ of clean images and rewards for query $i$. Appendix~\ref{app:implementation} specifies the observation sets and rollout schedules.

\paragraph{Centered clean-image regression.}
We form clean-image and reward contrasts using rollout-dependent centers. Trajectory banks use the query's own ODE completion and its reward; independent-seed groups use leave-one-out group means (Section~\ref{sec:setup}). Denoting these centers by $\widehat{\vect{m}}_i$ and $\widehat{\bar r}_i$, we define
\begin{equation}
 \vect{d}_{ij}=\vect{z}_j-\widehat{\vect{m}}_i,\qquad b_{ij}=r_j-\widehat{\bar r}_i,\qquad j\in\mathcal B_i.
 \label{eq:empirical_centering}
\end{equation}
For each nonzero displacement, divide the feature and label by the same length $\ell_{ij}=\|\vect{d}_{ij}\|$. Stack $\vect{d}_{ij}^{\top}/\ell_{ij}$ in $Q_i$ and $b_{ij}/\ell_{ij}$ in $\vect{b}_i$. The coefficient and its scaled regression direction are
\begin{equation}
 \widehat{\vect{u}}_i=\operatorname*{arg\,min}_{\vect{u}}\left\{\|Q_i\vect{u}-\vect{b}_i\|^2+\lambda_{\mathrm{sum}}\|\vect{u}\|^2\right\},
 \qquad \widehat{\vect{g}}_{i,t}=\kappa_t\widehat{\vect{u}}_i,
 \label{eq:query_fit}
\end{equation}
with $\lambda_{\mathrm{sum}}=0.3$ in the reported experiments. A small solve in the number of observations gives
\begin{equation}
 \widehat{\vect{u}}_i=Q_i^\top(Q_iQ_i^\top+\lambda_{\mathrm{sum}}I)^{-1}\vect{b}_i.
 \label{eq:query_dual}
\end{equation}
Centers are computed before removing zero-displacement rows; an empty fit returns zero. Constant rewards also give zero direction. Normalization and ridge stabilize the finite fit; we do not normalize the fitted direction again.

\paragraph{Reward-dependent query weights.}
Weights $a_i$ act on the policy update after fitting. Trajectory banks use $a_i=1$; the GenEval recipe selects the six lowest-reward queries per prompt. All other clean images, including full-score images, remain regression observations. Section~\ref{sec:row_ablation} compares this rule with continuous saturation weights and other selections. Query weighting does not change the ridge coefficient.

\paragraph{Policy update without trajectory backpropagation.}
Following Eq.~\eqref{eq:gradient_matching_loss}, we insert the scaled regression direction $\chi_t\widehat{\vect{g}}_{i,t}$ into the value-gradient term, where $\chi_t=\sigma_t^2/\alpha_t^2$. With stopped query weights and a shared normalization $Z_a=\E_{\mu_{\bthetaold}}[a_i]>0$, this yields
\begin{equation}
 \mathcal L_\tau=\frac{1}{\sg[Z_a]}\E_{\mu_{\bthetaold}}\!\left[a_iw_t\left(
 2\chi_t\langle\delta\vect{v}_{\btheta},\sg[\widehat{\vect{g}}_{i,t}]\rangle
 +\frac{2\tau}{\nu_t^2}\|\delta\vect{v}_{\btheta}\|^2\right)\right].
 \label{eq:kl_target}
\end{equation}
Next, center the cross term at $\vect{v}_{\bthetaold}$, the frozen policy that collected the current rollout bank. Replacing $\vect{v}_{\mathrm{ref}}$ by $\sg[\vect{v}_{\bthetaold}]$ in this term changes the loss only by a parameter-independent quantity, so the gradient is identical for every frozen batch. Using $\widehat{\vect{g}}_{i,t}=\kappa_t\widehat{\vect{u}}_i$ and $\omega_t=w_t\chi_t\kappa_t$ yields the equivalent centered loss
\begin{equation}
\begin{aligned}
 \mathcal L_\tau^{\mathrm{ctr}}=\frac{1}{\sg[Z_a]}\E_{\mu_{\bthetaold}}\!\Big[ a_i\Big(&
 2\omega_t\langle\vect{v}_{\btheta}-\sg[\vect{v}_{\bthetaold}],\sg[\widehat{\vect{u}}_i]\rangle\\
 &+\frac{2\tau w_t}{\nu_t^2}\|\vect{v}_{\btheta}-\vect{v}_{\mathrm{ref}}\|^2\Big)\Big].
\end{aligned}
 \label{eq:linear_update}
\end{equation}
The centered cross term is zero samplewise at $\btheta=\bthetaold$, removing loss fluctuations caused by the fixed-reference offset at the collection policy. Its gradient is preserved exactly; the quadratic penalty remains measured from $\vect{v}_{\mathrm{ref}}$. Since the gradients agree samplewise, their covariance is unchanged. Only the velocity evaluation receives gradients: rollouts, rewards, and regression are detached, so no reward derivative or adjoint through denoising is required. At $\tau=0$ the quadratic term is omitted, including for ODE rollouts; positive trajectory KL requires a nonzero diffusion coefficient. In practice we take $\tau=0$ and set the time weight $\omega_t=t$, which we found to work best, normalized to unit mean over the sampling grid; Appendix~\ref{app:implementation} gives the discrete loss.

\section{Main experiments}
\label{sec:experiments}
\subsection{Setup and evaluation}
\label{sec:setup}
We fine-tune SD3.5-Medium \citep{esser2024sd3} at $512\times512$ with rank-$32$, scale-$64$ LoRA \citep{hu2021lora} and ten-step rollouts. PickScore \citep{kirstain2023pickapic} uses Pick-a-Pic training prompts. DiffusionNFT and FDFO are reproduced from their official code on the same $8\times$A800-40GB hardware. Compute is wall time multiplied by eight. Each rollout collection gives one parameter update through Algorithm~\ref{alg:centered_regression}; Appendix~\ref{app:implementation} gives the configurations. The two-seed PickScore comparison is reported in Appendix~\ref{app:seed}.

Both the single-reward and five-reward comparisons use evaluation protocols and datasets aligned with DiffusionNFT for direct comparison: GenEval \citep{ghosh2023geneval} uses $553$ prompts with four images each; OCR uses $1{,}016$ evaluated prompts from the released OCR test split with PaddleOCR scoring; and DrawBench \citep{saharia2022imagen} uses $200$ prompts with five images each for PickScore, ClipScore, HPSv2.1, Aesthetic, ImageReward, and UnifiedReward. All our models are trained and evaluated with CFG $=1$. Evaluation uses a first-order ODE sampler with $40$ steps, aligned with DiffusionNFT and Flow-GRPO. Published results are reported as released.

Each recipe defines the image and reward centers in Eq.~\eqref{eq:empirical_centering} from its rollout construction.

\paragraph{PickScore: samples along one trajectory.}
We collect $24\times24$ trajectory-bank rollouts with $\eta=0.03$ and fit each direction from clean completions along the same stochastic trajectory. The query state's own ODE completion and its reward serve as the image and reward centers. All query trajectories receive actor updates.

\paragraph{GenEval: samples across trajectories.}
We collect $K=24$ independent ODE trajectories for each of $48$ prompts and select the six lowest-reward trajectories for actor updates. Each selected query is fitted from the other $23$ trajectories, whose mean clean image and mean reward serve as the regression centers. The query itself is excluded because its own reward selected it among the lowest six. The fitted direction is reused along its trajectory.

\paragraph{OCR.}
OCR combines graded credit for partially correct text with discrete recognition changes, placing it between smooth PickScore and plateaued GenEval. Both sampling schemes are effective in the ablation in Section~\ref{sec:estimator_ablation}. The main OCR-only run uses $8\times12$ trajectory-bank rollouts with $\eta=0.5$; the group variant uses lowest-six actor selection and also supplies the OCR quality-floor run.

\subsection{Joint optimization of five rewards}
\label{sec:joint_main}
We train one model on the five rewards of DiffusionNFT's joint experiment---GenEval, OCR, PickScore, ClipScore, and HPSv2.1---and also evaluate Aesthetic, ImageReward, and UnifiedReward. Training alternates the two sampling schemes of Algorithm~\ref{alg:centered_regression}: group updates on mixed GenEval, OCR, and Pick-a-Pic prompts, and trajectory-bank updates on Pick-a-Pic prompts with the composite quality reward $q=\mathrm{PickScore}/26+\mathrm{ClipScore}+\mathrm{HPSv2.1}$. On GenEval and OCR prompts, $q$ enters through the group-relative floor of Eq.~\eqref{eq:floor}. Starting from SD3.5-Medium, the model uses $1{,}000$ updates and \oMultiGPUh{} GPU-hours (Appendix~\ref{app:joint}).

Table~\ref{tab:multi} gives the result. Our model matches DiffusionNFT's joint model on GenEval, exceeds it on OCR, PickScore, ClipScore, HPSv2.1, ImageReward, and UnifiedReward, and trails only on Aesthetic ($\oMultiAes$ vs.\ $6.01$). DiffusionNFT does not report GPU-hours for this model; measured per-iteration runtime of its official code on our hardware implies at least $2{,}000$ GPU-hours for the $1{,}600$ forty-step iterations in its described stages (its table lists $1.7$k), roughly five times our budget.

\begin{table}[ht]
\caption{Five-reward comparison on the metric panel of DiffusionNFT Table~1. Gray marks the rewards used to train each model. Baseline rows are published results and retain their published inference protocols; $^{\dagger}$ denotes official checkpoints. DiffusionNFT lists $1.7$k iterations for its five-stage joint model, whose described stages total $1{,}600$ (Appendix~\ref{app:joint}); ours uses $1{,}000$ updates (\oMultiGPUh{} GPU-hours). Bold marks the best value in each column.}
\label{tab:multi}
\centering\small
\setlength{\tabcolsep}{3pt}
\resizebox{\linewidth}{!}{\begin{tabular}{lrcccccccc}
\toprule
Model & Iter. & GenEval & OCR & PickScore & ClipScore & HPSv2.1 & Aesthetic & ImgRwd & UniRwd\\
\midrule
SD3.5-M & -- & 0.63 & 0.59 & 22.34 & 0.285 & 0.279 & 5.36 & 0.85 & 3.03\\
\ \ + FlowGRPO$^{\dagger}$ (GenEval) & $>$5k & \cellcolor{gray!15}\textbf{0.95} & 0.66 & 22.51 & 0.293 & 0.274 & 5.32 & 1.06 & 3.18\\
\ \ + FlowGRPO$^{\dagger}$ (OCR) & 2k & 0.66 & \cellcolor{gray!15}\textbf{0.92} & 22.41 & 0.290 & 0.280 & 5.32 & 0.95 & 3.15\\
\ \ + FlowGRPO$^{\dagger}$ (PickScore) & 4k & 0.54 & 0.68 & \cellcolor{gray!15}23.50 & 0.280 & 0.316 & 5.90 & 1.29 & 3.37\\
\ \ + DiffusionNFT (joint) & 1.7k & \cellcolor{gray!15}0.94 & \cellcolor{gray!15}0.91 & \cellcolor{gray!15}23.80 & \cellcolor{gray!15}0.293 & \cellcolor{gray!15}0.331 & \textbf{6.01} & 1.49 & 3.49\\
\ \ + Ours (joint) & 1.0k & \cellcolor{gray!15}\oMultiGE & \cellcolor{gray!15}\textbf{\oMultiOCR} & \cellcolor{gray!15}\textbf{\oMultiPick} & \cellcolor{gray!15}\textbf{\oMultiClip} & \cellcolor{gray!15}\textbf{\oMultiHPS} & \oMultiAes & \textbf{\oMultiImg} & \textbf{\oMultiUni}\\
\bottomrule
\end{tabular}}
\end{table}

\subsection{Single-reward results and learning efficiency}
\label{sec:efficiency}
\label{sec:panels}
Table~\ref{tab:single} reports all eight metrics for the single-reward checkpoints and a separate OCR quality-floor run. The single-reward models reach GenEval $0.98$, OCR $0.98$, and PickScore $23.99$. The quality-floor run uses a composite training reward and is excluded from the single-reward speed comparison.

\begin{table}[t]
\caption{Single-reward results on the panel of DiffusionNFT Table~2, with a separate OCR quality-floor run. Gray marks the primary training reward; blue marks the three floor components (PickScore, ClipScore, HPSv2.1). Published rows retain their original protocols. Metric precision matches Table~\ref{tab:multi}: two decimals for GenEval, OCR, PickScore, Aesthetic, ImageReward, and UnifiedReward; three for ClipScore and HPSv2.1.}
\label{tab:single}
\centering\small
\setlength{\tabcolsep}{3pt}
\resizebox{\linewidth}{!}{\begin{tabular}{llrcccccccc}
\toprule
Model & Reward & Iter.\ (GPU-h) & GenEval & OCR & PickScore & ClipScore & HPSv2.1 & Aesthetic & ImgRwd & UniRwd\\
\midrule
SD3.5-M & -- & -- & 0.63 & 0.59 & 22.34 & 0.285 & 0.279 & 5.36 & 0.85 & 3.03\\
\midrule
FlowGRPO & GenEval & 4k & \cellcolor{gray!15}0.97 & 0.30 & 21.78 & 0.277 & 0.248 & 5.15 & 0.74 & 2.87\\
FlowGRPO & OCR & 1k & 0.66 & \cellcolor{gray!15}0.96 & 21.94 & 0.280 & 0.257 & 5.18 & 0.31 & 2.86\\
FlowGRPO & PickScore & 4k & 0.54 & 0.60 & \cellcolor{gray!15}23.62 & 0.257 & 0.295 & 6.42 & 1.17 & 3.17\\
\midrule
DiffusionNFT & GenEval & 1k & \cellcolor{gray!15}0.98 & 0.36 & 21.92 & 0.271 & 0.251 & 5.33 & 0.68 & 2.91\\
DiffusionNFT & OCR & 150 & 0.54 & \cellcolor{gray!15}0.97 & 21.63 & 0.281 & 0.246 & 5.19 & 0.37 & 2.81\\
DiffusionNFT & PickScore & 2k & 0.53 & 0.64 & \cellcolor{gray!15}24.03 & 0.270 & 0.315 & 6.17 & 1.29 & 3.40\\
\midrule
DiffusionNFT (repr.) & GenEval & 210 (137) & \cellcolor{gray!15}\nGEGE & \nGEOCR & \nGEPick & \nGEClip & \nGEHPS & \nGEAes & \nGEImg & \nGEUni\\
DiffusionNFT (repr.) & OCR & 50 (28) & \nOCRGE & \cellcolor{gray!15}\nOCROCR & \nOCRPick & \nOCRClip & \nOCRHPS & \nOCRAes & \nOCRImg & \nOCRUni\\
DiffusionNFT (repr.) & PickScore & 260 (132) & \nPickGE & \nPickOCR & \cellcolor{gray!15}\nPickPick & \nPickClip & \nPickHPS & \nPickAes & \nPickImg & \nPickUni\\
\midrule
Ours & GenEval & 400 (129) & \cellcolor{gray!15}\oGEGE & \oGEOCR & \oGEPick & \oGEClip & \oGEHPS & \oGEAes & \oGEImg & \oGEUni\\
Ours & OCR & 120 (19.5) & \oOCRGE & \cellcolor{gray!15}\oOCROCR & \oOCRPick & \oOCRClip & \oOCRHPS & \oOCRAes & \oOCRImg & \oOCRUni\\
Ours & PickScore & 200 (110) & \oPickGE & \oPickOCR & \cellcolor{gray!15}\oPickPick & \oPickClip & \oPickHPS & \oPickAes & \oPickImg & \oPickUni\\
\midrule
Ours & OCR + quality floor & 150 (41) & \oOCRfGE & \cellcolor{gray!15}\oOCRfOCR & \cellcolor{blue!15}\oOCRfPick & \cellcolor{blue!15}\oOCRfClip & \cellcolor{blue!15}\oOCRfHPS & \oOCRfAes & \oOCRfImg & \oOCRfUni\\
\bottomrule
\end{tabular}}
\end{table}

\begin{figure}[t]
\centering
\includegraphics[width=0.92\linewidth]{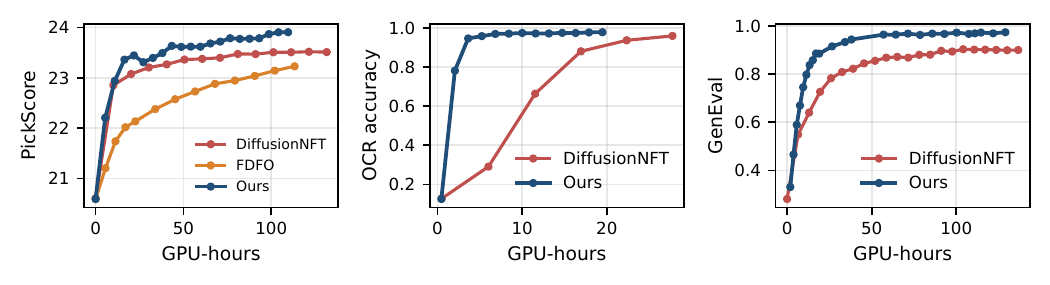}
\caption{Single-reward learning on $8\times$A800 hardware: each curve optimizes the named reward. All curves use the aligned DrawBench $200\times1$, OCR $1{,}016$, and official GenEval $553\times4$ datasets. FDFO appears only in the PickScore panel because FDFO does not evaluate the non-continuous OCR and GenEval rewards.}
\label{fig:main}
\end{figure}

Figure~\ref{fig:main} and Table~\ref{tab:efficiency} compare single-reward training time to matched reward thresholds, with each model optimized for the reward in its row. Clean-image regression is $2.1$--$2.6\times$ faster than DiffusionNFT on PickScore, $6.2$--$6.7\times$ on OCR, and $2.6$--$4.5\times$ on GenEval.

FDFO \citep{mcallister2026fdfo} does not evaluate discrete rewards such as OCR and GenEval, so Figure~\ref{fig:main} compares it on PickScore only. Our reward-adaptive rollouts address such discrete rewards directly (Section~\ref{sec:method}). Even on the continuous PickScore reward, FDFO needs $109.3$ GPU-hours to reach $23.20$, versus $14.3$ for our method, and stays below $23.50$ throughout its $113$-GPU-hour run.

\begin{table}[t]
\caption{Measured GPU-hours for single-reward training to shared reward thresholds in Figure~\ref{fig:main}.}
\label{tab:efficiency}
\centering\small
\setlength{\tabcolsep}{4pt}
\begin{tabular}{llrrr}
\toprule
Reward & Target & Ours & DiffusionNFT & Speed-up\\
\midrule
PickScore & 23.20 & 14.3 & 29.7 & $2.1\times$\\
& 23.50 & 38.2 & 98.5 & $2.6\times$\\
OCR & 0.80 & 2.2 & 14.9 & $6.7\times$\\
& 0.95 & 4.1 & 25.6 & $6.2\times$\\
GenEval & 0.80 & 11.5 & 30.3 & $2.6\times$\\
& 0.90 & 22.8 & 102.0 & $4.5\times$\\
\bottomrule
\end{tabular}
\end{table}

\paragraph{Image quality beyond the trained reward.}
\label{sec:hacking}
OCR-only training reaches its scalar reward quickly but often replaces requested scenes with large text on a flat background. After observing this reward hacking, we ran a separate, scheduled, group-relative quality-floor intervention. It reaches OCR $\oOCRfOCR$ and DrawBench PickScore $\oOCRfPick$ after $150$ updates ($41$ GPU-hours) while retaining scenes in the inspected samples (Table~\ref{tab:single}). Its training reward also uses PickScore, ClipScore, and HPSv2.1. Appendix~\ref{app:quality_floor} gives the reward formulation, schedule, failed alternatives, and qualitative comparisons.

At comparable trained-reward scores, our GenEval and PickScore checkpoints are numerically higher than the published DiffusionNFT single-reward rows on $6/7$ auxiliary metrics each (Table~\ref{tab:single}). For the OCR quality-floor checkpoint, GenEval, Aesthetic, ImageReward, and UnifiedReward did not enter its training reward; these scores and the inspected images complement the floor-component scores.

\section{Ablation Study}
\label{sec:ablation_study}
We test directional alignment, direction transport, reward contrasts, and actor weights. Appendix~\ref{app:ablations} gives the detailed results.

\subsection{Clean-image versus noisy-state regression: directional alignment}
\label{sec:endpoint_angle}
We compare noisy-state and clean-image fits using identical PickScore perturbations and reward differences, with the gradient through the numerical ODE as a reference (Appendix~\ref{app:endpoint_angle}). At a $200$-update checkpoint, direct clean-image regression raises the time-weighted mean signed cosine from $0.009$ to $0.107$: a paired gain of $0.098$ (95\% prompt-bootstrap CI $[0.092,0.106]$). The initial model shows the same trend (gain $0.204$).

\begin{figure}[t]
\centering
\includegraphics[width=\linewidth]{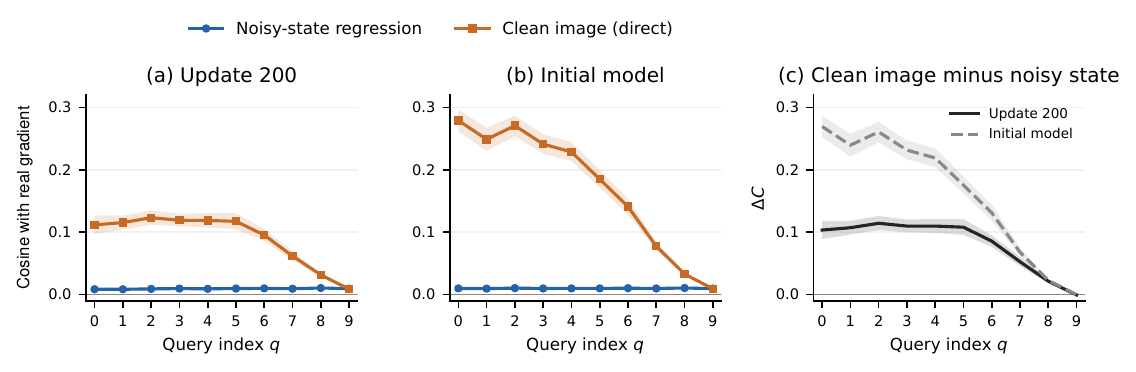}
\caption{PickScore cosine with the real gradient through the numerical ODE. Left and middle: noisy-state and direct clean-image regression. Right: paired clean-image-minus-noisy-state cosine. Shading denotes 95\% confidence intervals from resampling prompts.}
\label{fig:endpoint_angle}
\end{figure}

\subsection{Direct clean-image directions versus adjoint transport}
\label{sec:transport_ablation}
\begin{figure}[t]
\centering
\includegraphics[width=\linewidth]{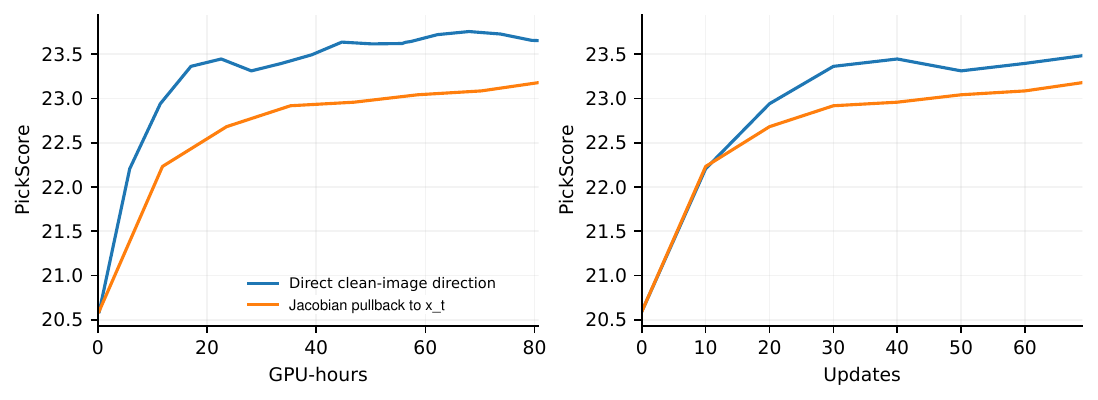}
\caption{PickScore training with direct clean-image directions or their norm-matched Jacobian pullback to $\vect{x}_t$, shown against compute (left) and updates (right). Both arms use the same trajectory bank and actor settings.}
\label{fig:adjoint_training}
\end{figure}

Keeping the PickScore recipe fixed, we replace each direct clean-image direction by its numerical ODE suffix pullback along the sample's own completion (Appendix~\ref{app:adjoint_transport}). We rescale the pullback to the original direction's norm for every sample and step, so transport changes only the direction and not the per-step update magnitudes. At $81$ GPU-hours ($69$ updates), norm-matched transport reaches $23.18$, versus $23.65$ for direct transfer at equal compute and $23.48$ at equal updates, and each transported update costs $2.1\times$ as much (Figure~\ref{fig:adjoint_training}; Appendix~\ref{app:adjoint_ablation}).

\subsection{Rollout bandwidth depends on reward continuity}
\label{sec:estimator_ablation}
\begin{figure}[t]
\centering
\includegraphics[width=\linewidth]{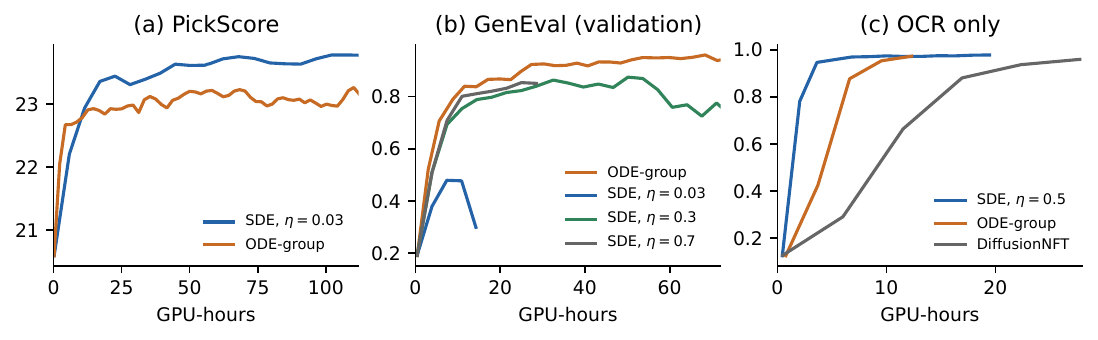}
\caption{\textbf{Rollout comparison across three rewards.} Training curves for trajectory-bank and independent-seed group regression. Batch sizes differ in GenEval and OCR, so these panels compare practical recipes at equal compute.}
\label{fig:estimator_reward}
\end{figure}
\paragraph{PickScore: local contrasts outperform independent-seed rollouts.}
Holding the $24\times24$ batch and optimizer fixed, we replace the small-noise bank by ODE-group regression, train all $24$ query trajectories, and use batch-standard-deviation reward scaling for the group arm. At $20$, $56$, and $112$ GPU-hours, the bank scores $23.41$, $23.63$, and $23.78$, versus $22.89$, $23.21$, and $23.18$ for the group arm. The latter remains $0.4$--$0.7$ behind at the matched budgets from $20$ GPU-hours onward. This is consistent with the local approximation in Eq.~\eqref{eq:curvature}: independent-seed rollouts give useful diversity but much longer clean-image differences.

\paragraph{GenEval: increasing local noise does not close the gap.}
We compare the $48\times24$ group recipe with all-state SDE banks using $48\times6$ trajectories at $\eta\in\{0.03,0.3,0.7\}$. At $28$ GPU-hours, group regression reaches $0.92$, versus $0.84$ and $0.85$ for the two larger noise levels; $\eta=0.03$ collapses to about $0.30$ by update $40$. Extending $\eta=0.3$ to $72$ GPU-hours gives $0.76$, versus $0.94$ for the group recipe. The group recipe leads at both reported budgets.

\paragraph{OCR.}
At update $40$, the trajectory bank and ODE-group both reach $0.97$, using $6.8$ and $12.3$ GPU-hours, respectively. Section~\ref{sec:hacking} reports the image-quality outcomes and the quality-floor extension.

\subsection{Effect of reward-adaptive weights on training}
\label{sec:row_ablation}
Keeping the GenEval group regression fixed, we vary the stopped actor weights in Eq.~\eqref{eq:actor}. Table~\ref{tab:row_weights} compares the resulting training scores at equal compute. Selecting the lowest six rewards gives the best score at $28$ GPU-hours; Appendix~\ref{app:row_weights} provides the full curves.

\begin{table}[!t]
\caption{GenEval actor-weight ablation. Scores are interpolated within the recorded validation curves at equal GPU-hours and rounded to two decimals; bold marks the best unrounded value in each column. Only the selected query trajectories receive actor gradients; all sibling clean images remain regression observations.}
\label{tab:row_weights}
\centering\small
\setlength{\tabcolsep}{8pt}
\begin{tabular}{lrrr}
\toprule
Stopped weight / selection & 10 GPU-h & 20 GPU-h & 28 GPU-h\\
\midrule
$a_i=1$ (all 24) & 0.72 & 0.84 & 0.86\\
$a_i=1-r_i$ & 0.71 & 0.88 & 0.89\\
$a_i=\mathbf{1}\{r_i<1\}$ & 0.74 & \textbf{0.88} & 0.88\\
Random 6 & 0.81 & 0.86 & 0.87\\
Highest 6 & 0.62 & 0.67 & 0.75\\
Lowest 6 (ours) & \textbf{0.81} & 0.87 & \textbf{0.92}\\
\bottomrule
\end{tabular}
\end{table}

\section{Conclusion}
\method uses centered clean-image regression to construct useful RL directions without adjoint backpropagation. Gradient matching identifies the update target, while population and directional analyses motivate regression in clean-image coordinates. Reward-adaptive rollouts and weights handle continuous and discrete scores within one regression method. A joint model matches or exceeds DiffusionNFT on seven of eight metrics at about one fifth of its estimated training compute; single-reward runs reach matched thresholds $2.1$--$6.7\times$ faster on identical hardware.

\clearpage

\subsection*{AI use statement}
\label{sec:statements}
Generative AI tools assisted drafting, translation, literature organization, LaTeX preparation, and plotting saved diagnostics. All reported numbers come from recorded evaluation files.

\subsection*{Reproducibility statement}
Sections~\ref{sec:theory} and~\ref{sec:method} define the population theory and evaluated estimator. Appendices~\ref{app:pathkl_control} and~\ref{app:preconditioning} give the control and regression proofs; Appendix~\ref{app:control_proof} contains the supplementary conditional sampling analysis. Appendices~\ref{app:ablations}--\ref{app:measurement} provide ablations, implementation details, and evaluation protocols; Appendix~\ref{app:hacking} documents reward exploitation. Our code is available at \url{https://github.com/jy-wang-tsinghua/DIFFUSIONRVGE}, and checkpoints will be released.

\clearpage
\appendix

\section{Proof of the trajectory-KL control identity}
\label{app:pathkl_control}
\paragraph{Assumptions and diffusion parameterization.}
Assume well-posed reference and controlled diffusions with the same initial law and diffusion coefficient, finite expected reward and control energy, and Girsanov integrability. The conditional exponential moment is positive and finite; its soft value is regular enough for It\^o's formula, with a zero-mean martingale term. Singular endpoints are treated by limits of truncated intervals with uniform integrability of the displayed terms.

For the Gaussian noising path, set $A_t=\dot\alpha_t/\alpha_t$ and $\nu_t^2=\partial_t\sigma_t^2-2A_t\sigma_t^2>0$ on the interior. The linear forward SDE $\dd\vect{X}_t=A_t\vect{X}_t\dd t+\nu_t\dd\vect{W}_t$ has the specified conditional Gaussian kernels. Its time reversal, initialized at the forward terminal law, has drift
\begin{equation*}
 \vect{b}_{\mathrm{ref}}=-A_t\vect{x}+\nu_t^2\nabla\log p_t=A_t\vect{x}-2\vect{v}_{\mathrm{ref}},\qquad
 \vect{v}_{\mathrm{ref}}=A_t\vect{x}-\tfrac12\nu_t^2\nabla\log p_t.
\end{equation*}
Its conditional clean law is the Gaussian noising posterior. The parameterization $\vect{b}_{\btheta}=A_t\vect{x}-2\vect{v}_{\btheta}$ therefore gives $\vect{u}_{\btheta}=-2\delta\vect{v}_{\btheta}$ and $\gamma_s^2=\nu_{T-s}^2$.

\paragraph{Proof of Theorem~\ref{thm:pathkl}.}
Write $\mathcal W_s=W_{T-s}$ and $\mathcal H_s=H_{T-s}$. The reference backward equation for $\mathcal H_s=e^{\mathcal W_s/\tau}$ yields
\begin{equation}
 (\partial_s+\mathcal L_s^{\mathrm{ref}})\mathcal W_s+\frac{\gamma_s^2}{2\tau}\|\nabla\mathcal W_s\|^2=0,\qquad
 \mathcal W_T=R,\qquad
 \mathcal L_s^{\mathrm{ref}}=\vect{b}_{\mathrm{ref}}\cdot\nabla+\tfrac12\gamma_s^2\Delta.
 \label{eq:hjb}
\end{equation}
Girsanov's identity and It\^o's formula under the controlled process give
\begin{equation}
\begin{aligned}
 \operatorname{KL}(P_{\btheta}\Vert P_{\mathrm{ref}})&=\frac12\E_{P_{\btheta}}\int_0^T\gamma_s^{-2}\|\vect{u}_{\btheta}\|^2\dd s,\\
 \E_{P_{\btheta}}R(\vect{Y}_T)-C_\tau&=\E_{P_{\btheta}}\int_0^T\left[\vect{u}_{\btheta}^\top\nabla\mathcal W_s-\frac{\gamma_s^2}{2\tau}\|\nabla\mathcal W_s\|^2\right]\dd s.
\end{aligned}
 \label{eq:girsanov_ito}
\end{equation}
Subtracting $\tau$ times the first identity from the second and completing the square proves Eq.~\eqref{eq:control_matching}. The maximizing correction is $\vect{u}_s^*=\gamma_s^2\nabla\mathcal W_s/\tau$ when admissible. Substituting $\vect{u}_{\btheta}=-2\delta\vect{v}_{\btheta}$ proves Eq.~\eqref{eq:velocity_matching_main}, including its positive residual sign. Appendix~\ref{app:alltau} gives the separate first-variation argument for the frozen-policy training loss.

The equivalent velocity identity is
\begin{equation}
 C_\tau-J_\tau=2\tau\E_{P_{\btheta}}\int_0^T
 \frac{\|\delta\vect{v}_{\btheta}+\gamma_s^2\nabla W_{T-s}/(2\tau)\|^2}{\gamma_s^2}\dd s.
 \label{eq:velocity_matching_main}
\end{equation}

\paragraph{Reward tilt and initialization.}
The path measure $\dd Q^*/\dd P_{\mathrm{ref}}=e^{R(\vect{Y}_T)/\tau}/Z_\tau$ has terminal law $\pi^*\propto e^{R/\tau}\rho$. Substitution into relative entropy proves Eq.~\eqref{eq:intro_control}. Its initial law is $\mu^*(\dd\vect{x})=H_T(\vect{x})\mu(\dd\vect{x})/Z_\tau$. The oracle drift started from $\mu^*$ realizes $Q^*$. With an arbitrary fixed initial law $\mu$, the same drift instead realizes the conditional tilt $e^{R(\vect{Y}_T)/\tau}/H_T(\vect{Y}_0)$. The two coincide when $H_T$ is constant, as at the independent-noise entrance in the main text.

\subsection{A common first variation for all penalty strengths}
\label{app:alltau}
In addition to the shared initial law and policy-independent diffusion, assume that $F_t^{\bthetaold,\tau}$ is $C^{1,2}$ on interior intervals, the following It\^o terms are integrable, and differentiation of the objective and endpoint limits are justified. For $\tau>0$ assume Girsanov integrability and finite expected control energy; no exponential moment or soft-value limit is needed for the first variation at $\tau=0$.

Write $\mathcal F_s=F_{T-s}^{\bthetaold,\tau}$ and $c_{\btheta,s}=\tau\|\vect{u}_{\btheta}\|^2/(2\gamma_s^2)$, taking $c_{\btheta,s}=0$ when $\tau=0$. Its backward equation is
\begin{equation*}
 (\partial_s+\mathcal L_s^{\bthetaold})\mathcal F_s=c_{\bthetaold,s},\qquad
 \mathcal F_T=R,\qquad
 \mathcal L_s^{\bthetaold}=\vect{b}_{\bthetaold}\cdot\nabla+\tfrac12\gamma_s^2\Delta.
\end{equation*}
It\^o's formula under $P_{\btheta}$ gives the exact performance difference
\begin{equation}
 J_\tau(\btheta)-J_\tau(\bthetaold)
 =\E_{P_{\btheta}}\int_0^T\!\left[
 \langle\vect{u}_{\btheta}-\vect{u}_{\bthetaold},\nabla\mathcal F_s\rangle
 -c_{\btheta,s}+c_{\bthetaold,s}\right]\dd s.
 \label{eq:alltau_difference}
\end{equation}
The integrand vanishes at $\btheta=\bthetaold$, so differentiating there leaves no first-order occupancy term. Consequently,
\begin{equation}
 \left.\nabla_{\btheta}J_\tau\right|_{\bthetaold}
 =\E_{P_{\bthetaold}}\int_0^T
 (\partial_{\btheta}\vect{u}_{\btheta}|_{\bthetaold})^\top
 \left(\nabla\mathcal F_s-\frac\tau{\gamma_s^2}\vect{u}_{\bthetaold}\right)\dd s.
 \label{eq:alltau_firstvariation}
\end{equation}
Substituting $\vect{u}_{\btheta}=-2\delta\vect{v}_{\btheta}$ proves the gradient identity for Eq.~\eqref{eq:gradient_matching_loss}, with $w_t=1/\varrho(t)$ when $t$ has sampling density $\varrho$. At zero penalty the cost terms are omitted before evaluation. For the reward-only ODE with generation drift $-\vect{v}_{\btheta}$, the same proof uses the transport equation and gives cross-term factor $1$ instead of $2$, assuming a differentiable continuation value. For $\tau>0$, $F_t^{\bthetaold,\tau}$ agrees with the optimal soft value $W_t$ at the oracle policy, linking the local update to Theorem~\ref{thm:pathkl}.

\paragraph{Regression uses the same return definition for all penalties.}
Perturb a query by $h\vecg{\epsilon}$ with $\E\vecg{\epsilon}=0$ and $M=\E[\vecg{\epsilon}\vecg{\epsilon}^{\top}]\succ0$. Let $\mathcal R_\tau$ be the terminal reward minus the sampled future frozen-policy control cost in Eq.~\eqref{eq:alltau_value}. Its conditional mean is $F_t^{\bthetaold,\tau}(\vect{x}+h\vecg{\epsilon})$. With a bounded Hessian on the perturbed states, finite second moments of $\mathcal R_\tau$ and $\E\|\vecg{\epsilon}\|^3<\infty$, the population regression satisfies
\begin{equation}
\begin{aligned}
 \vect{a}_{h,\tau}
 &=\operatorname*{arg\,min}_{\vect{a}}\E\!\left[
 (\mathcal R_\tau-\E\mathcal R_\tau-\vect{a}^\top h\vecg{\epsilon})^2\right]\\
 &=\frac1h M^{-1}\E[\vecg{\epsilon}F_t^{\bthetaold,\tau}(\vect{x}+h\vecg{\epsilon})]
 =\nabla F_t^{\bthetaold,\tau}(\vect{x})+O(h).
\end{aligned}
 \label{eq:alltau_regression}
\end{equation}
The normal equations give the middle equality; Taylor expansion gives the last. Thus the same regression principle applies for each fixed $\tau\geq0$. At zero penalty its labels are ordinary terminal rewards. For a fixed bank, centered feature matrix, row normalization and ridge coefficient, linearity of the fit also gives $\widehat{\vect{u}}_\tau=\widehat{\vect{u}}_R-\tau\widehat{\vect{u}}_C$, where the two coefficients fit terminal rewards and sampled future control energies.

\section{Covariance preconditioning and regression stability}
\label{app:preconditioning}
\paragraph{Covariance identities and proof of Proposition~\ref{thm:preconditioning}.}
For finite conditional moments and differentiation under the conditional integral, Bayes' rule gives
\begin{equation*}
 q_t(\dd\vect{z}\mid\vect{x})\propto e^{-\|\vect{x}-\alpha_t\vect{z}\|^2/(2\sigma_t^2)}\rho(\dd\vect{z}),\qquad
 \nabla_{\vect{x}}\log q_t(\vect{z}\mid\vect{x})=\kappa_t(\vect{z}-\vect{m}_t).
\end{equation*}
The derivative is of the density relative to $\rho$, so singular clean laws are allowed. Consequently $\nabla_{\vect{x}}\E[f(\vect{Z})\mid\vect{x}]=\kappa_t\Cov(\vect{Z},f(\vect{Z})\mid\vect{x})$. Taking $f=R$ proves Eq.~\eqref{eq:preconditioning}; taking $f=e^{R/\tau}$ yields
\begin{equation}
 \nabla_{\vect{x}}W_t=\tau\kappa_t\frac{\Cov(\vect{Z},e^{R/\tau}\mid\vect{x})}{H_t}.
 \label{eq:cov_soft}
\end{equation}
Neither identity differentiates the reward. The normal equations in Eq.~\eqref{eq:population_regression} give $\Sigma_t\vect{g}_t=\kappa_t\vect{c}_t$. If $\vect{a}\in\ker\Sigma_t$, then $\vect{a}^\top(\vect{Z}-\vect{m}_t)=0$ almost surely, so $\vect{a}^\top\vect{c}_t=0$. Thus $\vect{c}_t\in\operatorname{Range}(\Sigma_t)$ and the minimum-norm solution is $\vect{g}_t=\kappa_t\Sigma_t^\dagger\vect{c}_t$. Its inner product with $\nabla V_t$ is $(\nabla V_t)^\top\Sigma_t^\dagger\nabla V_t>0$ for nonzero $\nabla V_t$.

For an affine reward with slope $\vect{w}\in\operatorname{Range}(\Sigma_t)$, $\vect{c}_t=\Sigma_t\vect{w}$. The maps $\vect{w}\mapsto\nabla V_t$ and $\vect{w}\mapsto\vect{g}_t$ therefore have condition numbers $\kappa_+(\Sigma_t)$ and $1$ on that subspace, where $\kappa_+$ uses the nonzero eigenvalues.

\subsection{Directional advantage with mean-centered observations}
\label{app:posterior_dimension}
Let $P=UU^\top$ with $U^\top U=I_k$, $1\leq k<d$, and $\vect{Z}\sim\mathcal N(0,\sigma_{\mathrm{data}}^2P)$. For $R(\vect{z})=b+\vect{w}^\top\vect{z}$, $P\vect{w}=\vect{w}\neq0$, Gaussian conditioning gives
\begin{equation}
 \vect{m}_t(\vect{x})=a_tP\vect{x},\quad \Sigma_t=\lambda_tP,\quad
 a_t=\frac{\alpha_t\sigma_{\mathrm{data}}^2}{\alpha_t^2\sigma_{\mathrm{data}}^2+\sigma_t^2},\quad
 \lambda_t=\frac{\sigma_{\mathrm{data}}^2\sigma_t^2}{\alpha_t^2\sigma_{\mathrm{data}}^2+\sigma_t^2}.
 \label{eq:posterior_subspace}
\end{equation}
Since $W_t=V_t+\lambda_t\|\vect{w}\|^2/(2\tau)$, both gradients equal $a_t\vect{w}$, and $\vect{g}_t=\kappa_t\vect{w}$. Draw $m\geq2$ independent perturbations $\vecg{\delta}_j\sim\mathcal N(0,\eta^2I_d)$ and form a matrix $D$ with rows $\vecg{\delta}_j^\top$. Write $C=I_m-\vect{1}\vect{1}^\top/m$ for the centering matrix. The centered noisy rows, centered denoised means, and centered labels are
\begin{equation*}
 D_c=CD,\qquad Q_c=a_tD_cP,\qquad \vect{b}_c=a_tD_c\vect{w}=Q_c\vect{w}.
\end{equation*}
The unnormalized minimum-norm fits are
\begin{equation*}
 \widehat{\vect{h}}_{\mathrm{noisy}}=D_c^\dagger\vect{b}_c=a_t\Pi_S\vect{w},\qquad
 \widehat{\vect{g}}_{\mathrm{clean}}=\kappa_tQ_c^\dagger\vect{b}_c=\kappa_t\Pi_L\vect{w},
\end{equation*}
where $S=\row(D_c)$ and $L=\row(D_cP)$. If $B\in\R^{m\times(m-1)}$ is an orthonormal basis of $\vect{1}^\perp$, then $C=BB^\top$ and $B^\top D$ has $m-1$ independent isotropic Gaussian rows. Hence $S$ has the same distribution as the row space of an $(m-1)\times d$ Gaussian matrix. Rotational invariance gives
\begin{equation*}
 \E\Pi_S=\min((m-1)/d,1)I_d,\qquad
 \E\Pi_L=\min((m-1)/k,1)P.
\end{equation*}
For either nonzero projection, $\cos^2(\Pi_S\vect{w},\vect{w})=\|\Pi_S\vect{w}\|^2/\|\vect{w}\|^2$. Taking expectations proves Eq.~\eqref{eq:posterior_dimension}, including positive alignment almost surely.

\subsection{Ridge controls coefficient sensitivity}
\label{app:ridge_stability}
For the actual sum-loss convention in Eq.~\eqref{eq:query_fit}, let $Q=U\operatorname{diag}(s_j)V^\top$ be a singular value decomposition. Then
\begin{equation}
 \widehat{\vect{u}}=V\operatorname{diag}\!\left(\frac{s_j}{s_j^2+\lambda_{\mathrm{sum}}}\right)U^\top\vect{b},\qquad
 \|\widehat{\vect{u}}\|\leq\frac{\|\vect{b}\|}{2\sqrt{\lambda_{\mathrm{sum}}}}.
 \label{eq:ridge_stability}
\end{equation}
The bound follows from $s/(s^2+\lambda)\leq1/(2\sqrt\lambda)$. For fixed $Q$, perturbing the normalized labels by $\Delta\vect{b}$ changes the coefficient by at most $\|\Delta\vect{b}\|/(2\sqrt{\lambda_{\mathrm{sum}}})$. No rank assumption is required; this includes the linear dependence introduced by mean centering.

\paragraph{Local approximation and rollout radius.}
For $R$ with Hessian norm at most $L$ on a convex neighborhood containing the bank, set $\vect{d}_j=\vect{z}_j-\bar{\vect{z}}$ and $\vect{w}=\nabla R(\bar{\vect{z}})$. Taylor's theorem and $\sum_j\vect{d}_j=0$ give
\begin{equation}
 r_j-\bar r=\vect{w}^\top\vect{d}_j+e_j-\bar e,\qquad
 |e_j-\bar e|\leq\frac L2\left(\|\vect{d}_j\|^2+\frac1m\sum_k\|\vect{d}_k\|^2\right).
 \label{eq:curvature}
\end{equation}
This explains why smaller clean-image displacements can improve a local fit for smooth rewards.

\section{Supplementary analysis of Langevin-corrected sampling}
\label{app:control_proof}
This appendix provides a conditional consistency result for an idealized sampling dynamics at $\tau>0$. It assumes exact reference scores, ideal realization of the regression field, a vanishing score discrepancy, and a mixing inequality. Under these assumptions, the corrected dynamics tracks the reward-tilted target.

\paragraph{Idealized score approximation.}
\label{sec:consistency}
For $\tau>0$, to examine the structural effect of preconditioning, use $\chi_t=\sigma_t^2/\alpha_t^2$ and replace $\nabla W_t$ by $\chi_t\vect{g}_t$. In normalized coordinates $\vect{y}=\vect{x}/\alpha_t$, $\chi=e^{-2\ell}$, define $\vect{G}_\ell=\alpha_t\chi_t\vect{g}_t=\Sigma_\ell^\dagger\vect{c}_\ell$. Let $p_\ell=\rho*\mathcal N(0,\chi I)$ and $\pi_\ell^*=\pi^**\mathcal N(0,\chi I)$. The regression score is $\nabla\log p_\ell+\vect{G}_\ell/\tau$, with error
\begin{equation}
 \vect{d}_\ell=\frac{\vect{G}_\ell-\nabla_{\vect{y}}W_\ell}{\tau}
 =\frac{(I-\Sigma_\ell/\chi)\vect{G}_\ell+\nabla_{\vect{y}}V_\ell-\nabla_{\vect{y}}W_\ell}{\tau}.
 \label{eq:score_error}
\end{equation}
The two terms isolate covariance geometry and the ordinary-to-soft-value approximation. The following conditional result states when this preconditioning preserves the terminal target.

\begin{theorem}[Conditional consistency of idealized corrected sampling]
\label{thm:consistency}
Assume an exact reference score, ideal realization of the regression field, and the regularity conditions stated below. For a fixed $a_0>0$, let
\begin{equation}
 \dd\vect{Y}_\ell=(\chi+a_0)\left(\nabla\log p_\ell+\frac{\vect{G}_\ell}{\tau}\right)(\vect{Y}_\ell)\dd\ell+\sqrt{2a_0}\dd\vect{B}_\ell.
 \label{eq:corrected_dynamics}
\end{equation}
If $\pi_\ell^*$ satisfies a uniform log-Sobolev inequality, the initial relative entropy is finite, and $\E\|\vect{d}_\ell(\vect{Y}_\ell)\|^2\to0$, then
\begin{equation}
 \operatorname{KL}(\mu_\ell\Vert\pi_\ell^*)\to0,\qquad \mu_\ell\Rightarrow\pi^*\quad(\ell\to\infty).
 \label{eq:consistency_limit}
\end{equation}
\end{theorem}

\paragraph{Ideal control realization.}
The comparison isolates the regression approximation by assuming exact reference scores and
\begin{equation}
 \vect{v}_{\btheta}-\vect{v}_{\mathrm{ref}}=-\frac{\nu_t^2\chi_t}{2\tau}\vect{g}_t,\qquad
 \vect{s}_t^{\btheta}=\vect{s}_t^{\mathrm{ref}}+\frac{\chi_t\vect{g}_t}{\tau},\qquad
 \widetilde{\vect{u}}_t=\frac{\nu_t^2\chi_t}{\tau}\vect{g}_t.
 \label{eq:ideal_velocity}
\end{equation}
Here $\vect{s}_t^{\mathrm{ref}}=\nabla_{\vect{x}}\log p_t$ and $\vect{u}_t^*=\nu_t^2\nabla_{\vect{x}}W_t/\tau$. In normalized coordinates $\vect{y}=\vect{x}/\alpha_t$, write $P_\chi$ for Gaussian convolution of variance $\chi$. Reindex $V$, $W$, $\Sigma$ and $\vect{c}$ at $(\alpha_t\vect{y},t)$, and write $\vect{k}_\ell=\nabla_{\vect{y}}W_\ell/\tau=\nabla\log\pi_\ell^*-\nabla\log p_\ell$.

\paragraph{Scale and oracle control.}
Let $\chi_t=\sigma_t^2/\alpha_t^2$, $\vect{y}=\vect{x}/\alpha_t$, and $\chi=e^{-2\ell}$. Since $\alpha_t\chi_t\kappa_t=1$, the normalized regression field is $\vect{G}_\ell=\alpha_t\chi_t\vect{g}_t=\Sigma_\ell^\dagger\vect{c}_\ell$. The control targets satisfy the exact identity
\begin{equation}
 \widetilde{\vect{u}}_t=\chi_t\Sigma_t^\dagger\vect{u}_t^*
 +\frac{\nu_t^2\chi_t}{\tau}\Sigma_t^\dagger(\nabla_{\vect{x}}V_t-\nabla_{\vect{x}}W_t).
 \label{eq:control_residual}
\end{equation}
Thus scale-corrected preconditioning retains the ordinary/soft-value residual. Differentiating the conditional exponential moment gives
\begin{equation}
 \vect{k}_\ell=\frac{\Cov(\vect{Z},e^{R(\vect{Z})/\tau}\mid\vect{y})}{\chi\,\E[e^{R(\vect{Z})/\tau}\mid\vect{y}]},\qquad
 \nabla_{\vect{y}}V_\ell=\chi^{-1}\Sigma_\ell\vect{G}_\ell.
 \label{eq:normalized_scores}
\end{equation}
These identities yield the two-term discrepancy in Eq.~\eqref{eq:score_error}; neither residual is discarded.

\paragraph{Uniform endpoint assumptions.}
On a stable covariance support with orthogonal projector $P$, assume, uniformly in $\vect{y}$ as $\chi\downarrow0$,
\begin{equation}
\begin{gathered}
 \operatorname{Range}(\Sigma)=\operatorname{Range}(P),\qquad
 \Sigma=\chi P+O(\chi^2),\qquad
 \vect{c}=\chi\vect{w}+O(\chi^2),\qquad
 P\vect{w}=\vect{w},\\
 \frac{\Cov(\vect{Z},e^{R(\vect{Z})/\tau}\mid\vect{y})}{\E[e^{R(\vect{Z})/\tau}\mid\vect{y}]}
 =\tau^{-1}\chi\vect{w}+O(\chi^2),\qquad
 \sup_{\vect{y},\ell}\|\vect{w}\|<\infty.
\end{gathered}
 \label{eq:endpoint_moments}
\end{equation}
The support has fixed rank and the remainders are uniform in vector and operator norms. These assumptions on the conditional covariance suffice for the mean-square error criterion in Theorem~\ref{thm:consistency}; the noisy laws $p_\ell=P_\chi\rho$ and $\pi_\ell^*=P_\chi\pi^*$ have ambient densities for every $\chi>0$, even when $\rho$ is singular. In the Gaussian-subspace model of Eq.~\eqref{eq:posterior_subspace}, the covariance condition holds explicitly:
\begin{equation*}
 \Sigma_\ell=\frac{\chi}{1+\chi/\sigma_{\mathrm{data}}^2}P=\chi P-\frac{\chi^2}{\sigma_{\mathrm{data}}^2}P+O(\chi^3),\qquad
 \chi\Sigma_\ell^\dagger=(1+\chi/\sigma_{\mathrm{data}}^2)P.
\end{equation*}
On this support, $\lambda_{\min}^+(\Sigma)\geq\chi/2$ for sufficiently small $\chi$. Hence
\begin{equation*}
 \vect{G}_\ell-\vect{w}=\Sigma^\dagger(\vect{c}-\Sigma\vect{w})=O(\chi),\qquad
 \vect{k}_\ell=\tau^{-1}\vect{w}+O(\chi),\qquad
 \vect{d}_\ell=O(\chi).
\end{equation*}
This proves the scale-consistency assertion. In the original coordinates,
\begin{equation*}
 \frac{\tau}{\nu_t^2}\|\widetilde{\vect{u}}_t-\vect{u}_t^*\|=\|\chi_t\vect{g}_t-\nabla_{\vect{x}}W_t\|=\frac{\tau}{\alpha_t}\|\vect{d}_\ell\|=O(\chi_t).
\end{equation*}
An unnormalized control-drift bound additionally depends on the schedule $\nu_t^2$.

\paragraph{Scope and regularity of the conditional guarantee.}
Assume $\tau>0$, $0<Z_\tau=\int e^{R/\tau}\dd\rho<\infty$, well-posed corrected dynamics, positive smooth noisy densities, and finite initial relative entropy at a finite $\ell_0$. Assume the differentiations and integrations by parts below are justified, and the score-error energy is locally integrable. Reference scores and the displayed guidance field are realized exactly. For $\mathcal H_\ell=\operatorname{KL}(\mu_\ell\Vert\pi_\ell^*)$ and $\mathcal I_\ell=\E_{\mu_\ell}\|\nabla\log(\mu_\ell/\pi_\ell^*)\|^2$, assume $\mathcal I_\ell\ge2\kappa\mathcal H_\ell$ for some $\kappa>0$ independent of $\ell\ge\ell_0$.

\paragraph{Entropy proof.}
Gaussian convolution and $\chi=e^{-2\ell}$ give $\partial_\ell\pi_\ell^*=-\chi\Delta\pi_\ell^*$. Hence the oracle drift $(\chi+a_0)\nabla\log\pi_\ell^*$ with diffusion $\sqrt{2a_0}I$ exactly transports $\pi_\ell^*$. The approximate drift differs by $\vect D_\ell=(\chi+a_0)\vect d_\ell$. Writing $f_\ell=\log(\mu_\ell/\pi_\ell^*)$, the relative-entropy identity and Young's inequality imply
\begin{equation}
 \mathcal H_\ell'=-a_0\mathcal I_\ell+\E_{\mu_\ell}\langle\vect D_\ell,\nabla f_\ell\rangle
 \le -a_0\kappa\mathcal H_\ell+\varepsilon_\ell,
 \qquad \varepsilon_\ell=\frac{(\chi+a_0)^2}{2a_0}\E_{\mu_\ell}\|\vect d_\ell\|^2.
 \label{eq:entropy_rate}
\end{equation}
Consequently,
\begin{equation}
 \mathcal H_\ell\le e^{-a_0\kappa(\ell-\ell_0)}\mathcal H_{\ell_0}
 +\int_{\ell_0}^\ell e^{-a_0\kappa(\ell-r)}\varepsilon_r\dd r.
 \label{eq:entropy_decay}
\end{equation}
Since $\varepsilon_\ell\to0$, split the integral at a sufficiently large fixed time to see that both terms vanish. This proves relative-entropy tracking. Pinsker's inequality and $\pi_\ell^*\Rightarrow\pi^*$ then give the weak endpoint limit. For a singular clean target, the endpoint conclusion is weak convergence, not total-variation convergence to that singular measure. More generally, the same proof needs only that the exponentially weighted error integral in Eq.~\eqref{eq:entropy_decay} tends to zero.

\paragraph{Discrete rewards and a smooth reference.}
For bounded measurable $R$, the positive-noise covariance identity requires no reward derivative, and the convergence theorem applies to an explicitly specified bounded smooth reference $R_\varepsilon$ at fixed $\varepsilon>0$, whenever its assumptions hold. If $R_\varepsilon\to R$ in $L^1(\rho)$ with a common bound, then
\begin{equation*}
 \pi^{*,\varepsilon}(\dd\vect{z})=\frac{e^{R_\varepsilon(\vect{z})/\tau}\rho(\dd\vect{z})}{\int e^{R_\varepsilon/\tau}\dd\rho}
 \longrightarrow\pi^*(\dd\vect{z})\quad\text{in total variation},
\end{equation*}
because the exponential is Lipschitz on the bounded reward range and the normalizers converge.

\section{Numerical adjoint used in the transport ablation}
\label{app:adjoint_transport}
Let $F_t$ be a deterministic ODE completion with differentiable velocity. Transporting a fitted clean-image direction $\vect{b}$ to the current state requires the Jacobian product $DF_t^\top\vect{b}$. For a smooth path, the continuous adjoint solves
\begin{equation}
 \dot{\vecg{\lambda}}_s=-D_{\vect{x}}\vect{v}(\vect{x}_s,s)^\top\vecg{\lambda}_s,\qquad
 \vecg{\lambda}_0=\vect{b},\qquad \vecg{\lambda}_t=DF_t(\vect{x}_t)^\top\vect{b}.
 \label{eq:adjoint_boundary}
\end{equation}
Indeed, a state perturbation satisfies $\dot{\vecg{\xi}}_s=D_{\vect{x}}\vect{v}\,\vecg{\xi}_s$, so $\vecg{\lambda}_s^\top\vecg{\xi}_s$ is constant. For an Euler completion $\vect{z}_{k+1}=\vect{z}_k+\Delta t_k\vect{v}(\vect{z}_k,t_k)$, define $J_k=I+\Delta t_kD_{\vect{x}}\vect{v}(\vect{z}_k,t_k)$. A velocity perturbation at step $q$ gives
\begin{equation}
 \delta R=\Delta t_q\langle\vecg{\lambda}_{q+1},\delta\vect{v}_q\rangle,\qquad
 \vecg{\lambda}_{q+1}=J_{q+1}^\top\cdots J_{N-1}^\top\vect{b},\qquad \vect{b}=\nabla R(\vect{z}_N).
 \label{eq:adjoint_actor_convention}
\end{equation}
The ablation replaces $\vect{b}$ by $\widehat{\vect{u}}_q$, evaluates this suffix along the query's own completion by vector--Jacobian products, and rescales it to the original coefficient's norm. An empty suffix is the identity. This keeps per-step update magnitudes fixed while testing the cost and benefit of transporting directions. For a coefficient error $\vect{e}$, the continuous pullback error is bounded by $\|DF_t\|_{\mathrm{op}}\|\vect{e}\|$. Direct transfer avoids these Jacobian products; its training performance is evaluated in Section~\ref{sec:transport_ablation}.

\section{Complete reward and weighting ablations}
\label{app:ablations}
Every training arm uses $8\times$A800-40GB. We show both compute and update count because the training choices change update cost. Budget-matched scores are linearly interpolated within the recorded curves.

\subsection{Reward-dependent actor weights}
\label{app:row_weights}
All six arms use the same mean-centered ODE-group clean-image regression and $48\times24$ collected trajectories. We change only the query weights: all $24$ ($a_i=1$); $a_i=1-r_i$; $a_i=\mathbf{1}\{r_i<1\}$; random six; highest six; lowest six. Random selection uses the first six independent-seed rollouts, so selection is independent of reward. The scalar functions consider all $24$ queries; zero-weight queries need neither a fit for their own actor nor a backward pass. Their clean images still serve as sibling observations. The denominator is the sum of actor weights, as in Eq.~\eqref{eq:actor}, not the number of candidates.

\begin{figure}[ht]
\centering
\includegraphics[width=\linewidth]{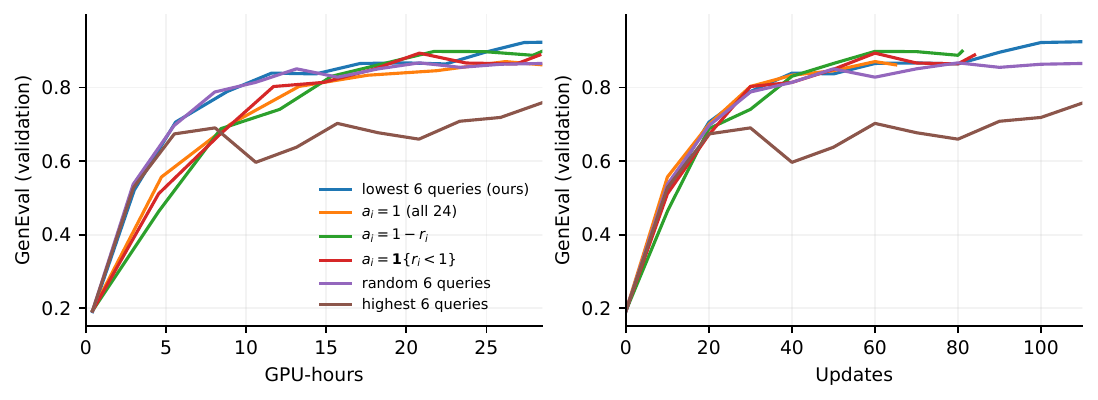}
\caption{GenEval query weights at a $28$-GPU-hour budget, shown against compute and updates.}
\label{fig:weights_full}
\end{figure}

At $28$ GPU-hours the two ceiling-aware arms have completed approximately $81$--$84$ updates, versus about $110$ for the lowest-six arm.

The saturation motivation also has a frozen-policy check: at an early checkpoint, a one-sided perturbation of $3\%$ of state RMS along the group fit changes the reward of failed samples by $+0.064$ and successful samples by $-0.019$, with random-direction controls around $\pm0.003$. During updates $28$--$65$, roughly $74\%$ of the all-24 arm's actor samples already have maximum reward.

\begin{figure}[ht]
\centering
\includegraphics[width=\linewidth]{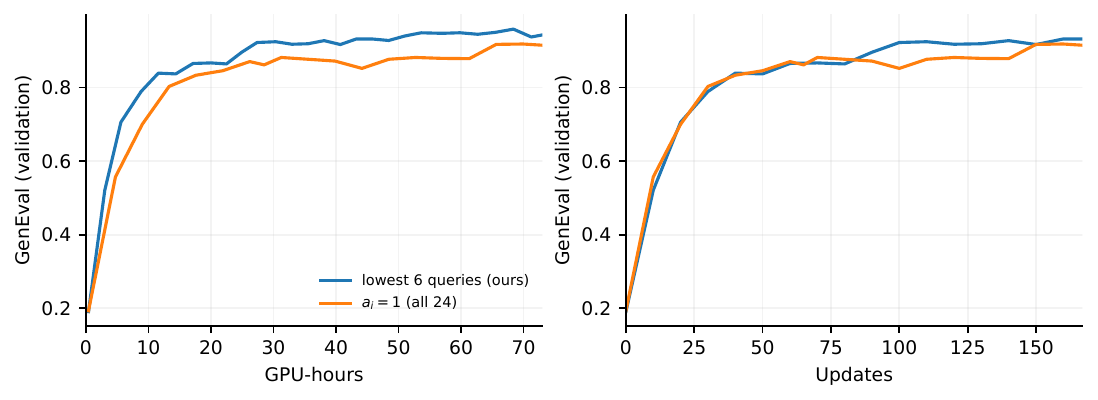}
\caption{The all-24 continuation. The lowest-six advantage persists beyond the initial budget but narrows near $72$ GPU-hours.}
\label{fig:all24_long}
\end{figure}

\begin{table}[ht]
\caption{Longer GenEval comparison at fixed compute, using interpolation within the saved curves.}
\centering\small
\begin{tabular}{lrrrrr}
\toprule
GPU-hours & 28 & 35 & 44 & 52 & 61\\
\midrule
Lowest 6 & 0.92 & 0.92 & 0.93 & 0.94 & 0.95\\
All 24 & 0.86 & 0.88 & 0.85 & 0.88 & 0.88\\
\bottomrule
\end{tabular}
\end{table}

\subsection{Norm-matched adjoint transport of the clean-image estimate}
\label{app:adjoint_ablation}
This arm keeps the PickScore $24\times24$ bank recipe and replaces each fitted clean-image direction by the suffix pullback in Eq.~\eqref{eq:adjoint_actor_convention}. The Jacobians are evaluated along each sample's own Euler completion from $\vect{z}_{q+1}=\vect{x}_q+\Delta t_q\vect{v}(\vect{x}_q,t_q)$, the completion that produced its clean image, and applied by vector--Jacobian products without a dense Jacobian. Each transported direction is then rescaled to the root-mean-square of the direction it replaces, separately for every sample and step, so the update keeps the recipe's magnitudes across samples and times and changes only the orientation. Over the $69$ updates, the transported direction has median cosine $0.18$ with the clean-image direction at the first step (range $0.15$--$0.41$), rising monotonically to $0.79$ at step $7$ and $1$ at the last two. The transport runs in the training precision (bfloat16); on eight probe samples its median cosine with a float32 reference is $0.71$, $0.95$, $0.97$, $0.99$, and $1.00$ at steps $0$, $2$, $4$, $6$, and $8$.

At $12$, $24$, $35$, $47$, $59$, $70$, and $81$ GPU-hours (updates $10,20,\dots,60$ and $69$), the transported arm scores $22.23$, $22.68$, $22.92$, $22.96$, $23.04$, $23.09$, and $23.18$. At the same compute the direct arm scores $22.97$, $23.42$, $23.43$, $23.63$, $23.66$, $23.75$, and $23.65$; at the same updates it scores $22.21$, $22.94$, $23.36$, $23.45$, $23.31$, $23.40$, and $23.48$. From update $20$ onward the per-update deficit ($0.26$--$0.49$) exceeds the difference between the two training seeds at matched updates (at most $0.14$ through update $70$; Appendix~\ref{app:seed}).

\subsection{PickScore directional-alignment probe}
\label{app:endpoint_angle}
We compare noisy-state and clean-image directions from matched PickScore perturbations at the initial model and an update-$200$ checkpoint. We measure their cosine with the real reward gradient $\vect{g}_{t_q}^{\mathrm{ODE}}=\nabla_{\vect{x}}(R\circ F_{t_q})(\vect{x}_q)$, differentiated through the numerical ODE completion, VAE, and PickScore (Figure~\ref{fig:endpoint_angle}).

We average signed cosines within prompts with time weights proportional to $t_q$ and use prompt-bootstrap resampling for confidence intervals. The clean-image direction has higher cosine with the real gradient, although its advantage diminishes near the last query (Figure~\ref{fig:endpoint_angle}).

Using the actual trajectory bank, the direct estimate also has positive cosine with $\vect{g}_{t_q}^{\mathrm{ODE}}$ at update $200$.

\subsection{PickScore training-seed replication}
\label{app:seed}
\begin{figure}[ht]
\centering
\includegraphics[width=\linewidth]{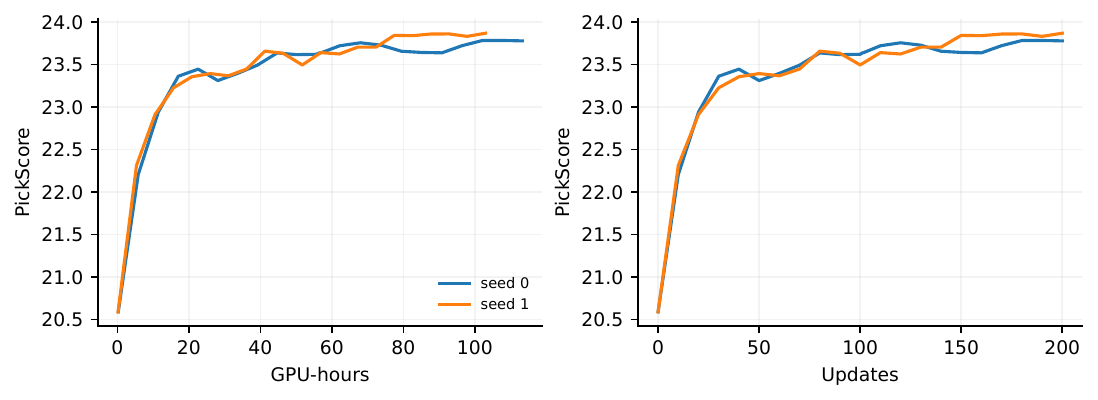}
\caption{PickScore training across seeds, shown against GPU-hours (left) and updates (right).}
\label{fig:pickscore_seeds}
\end{figure}

We test the PickScore training recipe with multiple seeds. Figure~\ref{fig:pickscore_seeds} shows their learning curves.

\section{Implementation and reproducibility protocol}
\label{app:implementation}
\paragraph{Reward-adaptive rollouts, one clean-image regression.}
Table~\ref{tab:config} gives the configurations for Algorithm~\ref{alg:centered_regression}. All use SD3.5-Medium, the ten-step shifted SD3 grid, CFG $=1$, and one parameter update per collection. The reference rows and update normalization for the sampling schemes in Section~\ref{sec:setup} are specified below.

\begin{table}[ht]
\caption{Configurations at the reported checkpoints. OCR floor uses Eq.~\eqref{eq:floor} with $\lambda_{\mathrm{floor}}=10$ and the margin schedule $m=0\to0.03\to0.06$ at updates $40$ and $100$.}
\label{tab:config}
\centering\small
\setlength{\tabcolsep}{4pt}
\resizebox{\linewidth}{!}{\begin{tabular}{lcccc}
\toprule
Setting & PickScore & OCR & GenEval & OCR floor\\
\midrule
Estimator & bank & bank & group & group\\
Prompts / update & 24 & 8 & 48 & 48\\
Trajectories / prompt & 24 & 12 & 24 & 24\\
Trained trajectories / prompt & 24 & 12 & 6 lowest & 6 lowest\\
Rollout & SDE, $\eta=0.03$ & SDE, $\eta=0.5$ & ODE & ODE\\
Reward evaluations / trajectory & 11 & 11 & 1 & 1\\
Regression rows & head-8 & head-8 & 23 siblings & 23 siblings\\
Ridge $\lambda_{\mathrm{sum}}$ & 0.3 & 0.3 & 0.3 & 0.3\\
Time weights & $\omega_q\propto t_q$ & $\omega_q\propto t_q$ & $\omega_q\propto t_q$ & $\omega_q\propto t_q$\\
Initial learning rate & $3\times10^{-4}$ & $3\times10^{-4}$ & $3\times10^{-4}$ & $3\times10^{-4}$\\
Reported update & 200 & 120 & 400 & 150\\
GPU-hours ($8\times$A800) & 110 & 19.5 & 129 & 41\\
\bottomrule
\end{tabular}}
\end{table}

\begin{algorithm}[ht]
\caption{One \method update with centered clean-image regression}
\label{alg:centered_regression}
\small
\begin{algorithmic}[1]
\Require Collection parameters $\bthetaold$, reward $R$, grid $t_0>\cdots>t_N=0$, ridge $\lambda_{\mathrm{sum}}$
\Require Rollout and query-weight rules (Section~\ref{sec:setup})
\State Sample prompts $c_i$ and collect query trajectories $\{\vect{x}_{i,q}\}_{q=0}^{N}$.
\State Complete rollout states to clean images $\vect{z}_j$ and evaluate $r_j=R(\vect{z}_j,c_j)$.
\State Assign actor weights $a_i$ from the collected rewards; set $Z_a=\sum_i a_i$.
\ForAll{query trajectories $i$ with $a_i>0$ and times $q=0,\ldots,N-1$}
  \State Form same-prompt references $\mathcal B_{i,q}$ and set the rollout-dependent image and reward centers $(\widehat{\vect{m}}_{i,q},\widehat{\bar r}_{i,q})$.
  \ForAll{$j\in\mathcal B_{i,q}$}
    \State $\vect{d}_{ij}\gets\vect{z}_j-\widehat{\vect{m}}_{i,q}$; $b_{ij}\gets r_j-\widehat{\bar r}_{i,q}$.
  \EndFor
  \State Discard zero displacements; set $\ell_{ij}=\|\vect{d}_{ij}\|_2$ for each retained row.
  \State Stack $\vect{d}_{ij}^{\top}/\ell_{ij}$ into $Q_{i,q}$ and $b_{ij}/\ell_{ij}$ into $\vect{b}_{i,q}$.
  \State $\widehat{\vect{u}}_{i,q}\gets Q_{i,q}^{\top}(Q_{i,q}Q_{i,q}^{\top}+\lambda_{\mathrm{sum}}I)^{-1}\vect{b}_{i,q}$; use $0$ for an empty fit.
\EndFor
\State Set $\omega_q=t_q/(N^{-1}\sum_{p=0}^{N-1}t_p)$; detach directions, weights, and $Z_a$.
\If{$Z_a>0$}
  \State $\delta\vect{v}_{\btheta,i,q}\gets\vect{v}_{\btheta}(\vect{x}_{i,q},t_q,c_i)-\sg[\vect{v}_{\bthetaold}(\vect{x}_{i,q},t_q,c_i)]$.
  \State $\mathcal L\gets\dfrac{2}{NZ_a}\sum_{i:a_i>0}\sum_{q=0}^{N-1}a_i\omega_q\langle\delta\vect{v}_{\btheta,i,q},\sg[\widehat{\vect{u}}_{i,q}]\rangle$.
  \State Starting from $\btheta=\bthetaold$, take one parameter update to minimize $\mathcal L$.
\EndIf
\end{algorithmic}
\end{algorithm}

\paragraph{Exploration transition.}
With $\Delta t_q=t_{q+1}-t_q<0$ and time-dependent diffusion coefficient $\gamma_q$, its interior-step discretization is
\begin{equation}
\begin{split}
 \vect{x}_{i,q+1}={}&\vect{x}_{i,q}+\Delta t_q\left[\vect{v}_{\bthetaold}(\vect{x}_{i,q},t_q,c)
 +\frac{\gamma_q^2}{2t_q}\widehat{\vecg{\epsilon}}_{i,q}\right]
 +\gamma_q\sqrt{-\Delta t_q}\,\vecg{\xi}_{i,q},\\
 &\widehat{\vecg{\epsilon}}_{i,q}=\vect{x}_{i,q}+(1-t_q)\vect{v}_{\bthetaold}(\vect{x}_{i,q},t_q,c),
 \qquad \vecg{\xi}_{i,q}\sim\mathcal N(0,I).
\end{split}
\label{eq:sampler}
\end{equation}

\paragraph{Noise coefficient and boundaries.}
The configuration's constant noise mode holds the radius parameter $\eta$ fixed, not the per-transition standard deviation. The recorded coefficient is
\begin{equation}
 \gamma_q^2=\frac{\eta^2t_q}{1-\widetilde t_q},\qquad
 \widetilde t_q=\begin{cases}t_1,&q=0,\\t_q,&q>0.\end{cases}
\end{equation}
The factor $\sqrt{-\Delta t_q}$ makes the perturbation step-dependent. Using $1-t_1$ at $t_0=1$ avoids the singular denominator.

\paragraph{Regression rows and solve.}
Equations~\eqref{eq:empirical_centering}--\eqref{eq:query_dual} state the fit for a generic query. Here $i$ indexes trajectories in a prompt group, $q$ indexes query time, and $j\in\mathcal B_{i,q}$ indexes reference completions; $(\widehat{\vect{m}}_{i,q},\widehat{\bar r}_{i,q})$ denote the centers in Eq.~\eqref{eq:empirical_centering} at query time $t_q$. For each retained nonzero displacement, define
\begin{equation}
 \vecg{\phi}_{i,q,j}=\frac{\vect{z}_j-\widehat{\vect{m}}_{i,q}}{\|\vect{z}_j-\widehat{\vect{m}}_{i,q}\|_2},\qquad
 \widetilde b_{i,q,j}=\frac{r_j-\widehat{\bar r}_{i,q}}{\|\vect{z}_j-\widehat{\vect{m}}_{i,q}\|_2}.
 \label{eq:rows}
\end{equation}
Stack $\vecg{\phi}_{i,q,j}^\top$ into $Q_{i,q}$ and the scalar targets $\widetilde b_{i,q,j}$ into $\vect{b}_{i,q}$. The fitted coefficient is
\begin{equation}
 \widehat{\vect{u}}_{i,q}
 =\operatorname*{arg\,min}_{\vect{u}}\left\{\|Q_{i,q}\vect{u}-\vect{b}_{i,q}\|_2^2+\lambda_{\mathrm{sum}}\|\vect{u}\|_2^2\right\}
 = Q_{i,q}^{\top}(Q_{i,q}Q_{i,q}^{\top}+\lambda_{\mathrm{sum}} I)^{-1}\vect{b}_{i,q}.
 \label{eq:ridge}
\end{equation}
This is Eq.~\eqref{eq:query_dual} with $\ell_{ij}=\|\vect{d}_{ij}\|$, before scaling the direction by $\kappa_t$; the sum-loss ridge coefficient is $\lambda_{\mathrm{sum}}=0.3$. The dual solve scales with the number of observations. The fitted coefficient $\widehat{\vect{u}}$ summarizes the observed reward changes across the chosen clean-image contrasts.

\paragraph{Trajectory-bank samples.}
For the trajectory bank, use $\mathcal B_{i,q}=\{(i,m):0\leq m<8,\ m\neq q\}$ on the ten-step grid, with
\begin{equation}
 \vect{z}_{i,q}=F_{t_q}^{\bthetaold}(\vect{x}_{i,q}),\qquad
 \widehat{\vect{m}}_{i,q}=\vect{z}_{i,q},\qquad
 \widehat{\bar r}_{i,q}=R(\vect{z}_{i,q}).
 \label{eq:pickscore_centers}
\end{equation}
The numerators in Eq.~\eqref{eq:rows} become $\vect{z}_{i,m}-\vect{z}_{i,q}$ and $r_{i,m}-r_{i,q}$. This head-8 reference set avoids the shortest late-stage chords. All retained completions enter one regression with the same rule.

\paragraph{Independent-seed group samples.}
\label{app:centering}
For independent-seed groups, use $\mathcal B_{i,q}=\{j:j\ne i\}$, with
\begin{equation}
 \widehat{\vect{m}}_{i,q}=\bar{\vect{z}}_{-i}
 =\frac{1}{K-1}\sum_{j\ne i}\vect{z}_j,\qquad
 \widehat{\bar r}_{i,q}=\bar r_{-i}
 =\frac{1}{K-1}\sum_{j\ne i}r_j.
 \label{eq:geneval_centers}
\end{equation}
The raw rows satisfy $\sum_{j\ne i}\vect{d}_{ij}=0$ and $\sum_{j\ne i}b_{ij}=0$. Normalize these rows as in Eq.~\eqref{eq:rows}, solve Eq.~\eqref{eq:ridge}, and reuse the resulting $\widehat{\vect{u}}_i$ at every query time on trajectory $i$.

\paragraph{Discrete velocity loss and normalization.}
The reported updates implement Eq.~\eqref{eq:linear_update} on the $N$-step grid. For $B$ collected trajectories with stopped query weights $a_i$ and $Z_a=\sum_{i=1}^Ba_i>0$,
\begin{equation}
 \mathcal L_{\mathrm{linear}}(\btheta;\bthetaold)
 =\frac{1}{N\sg[Z_a]}\sum_{i=1}^{B}\sum_{q=0}^{N-1}
 2\omega_q\sg[a_i]\left\langle
 \vect{v}_{\btheta}(\vect{x}_{i,q},t_q,c_i)-\sg[\vect{v}_{\bthetaold}(\vect{x}_{i,q},t_q,c_i)],
 \sg[\widehat{\vect{u}}_{i,q}]\right\rangle,
 \label{eq:actor}
\end{equation}
and the loss is zero when $Z_a=0$. Here $\omega_q=t_q/(N^{-1}\sum_pt_p)$ is the normalized empirical product $\omega_{t_q}=w_{t_q}\chi_{t_q}\kappa_{t_q}$ from Section~\ref{sec:method}. Equation~\eqref{eq:actor} is the discrete form of Eq.~\eqref{eq:linear_update}, with $t$ normalized to unit mean over the grid and the average taken over queries with $a_i>0$. We set $a_i=1$ for trajectory banks and $a_i=\mathbf 1\{i\in\operatorname{Lowest}_6(\vect{r}_{c_i})\}$ for GenEval, breaking ties randomly. Both binary selection and the continuous weights compared in Section~\ref{sec:row_ablation} act on the stopped actor term after the fit.

Both weights and denominator are stopped, and a zero denominator gives zero loss. Setting all $a_i=1$ recovers the unweighted batch average; binary lowest-six weights recover the selected-sample average exactly. Zero-weight queries require no actor backward pass. Every reported run uses these unit-mean linear time weights. All-equal rewards give a zero mean-centered group direction even if rank selection retains six queries. The stricter combination $\mathbf{1}\{r_i<1\}\mathbf{1}\{i\in\operatorname{Lowest}_6\}$ is distinct from the ungated lowest-six reference and is not silently substituted for it.

\subsection{Five-reward joint training}
\label{app:joint}
The joint model of Table~\ref{tab:multi} alternates the two update types above on DiffusionNFT's five training rewards. With $q=\mathrm{PickScore}/26+\mathrm{ClipScore}+\mathrm{HPSv2.1}$, Pick-a-Pic prompts receive $r=10q$, and GenEval and OCR prompts receive their task reward with the floor of Eq.~\eqref{eq:floor}, $r=R_{\mathrm{task}}+10\min(q-\bar q_c+m,0)$, where $m=0$ before update $700$ and $m=0.03$ afterwards. A group update draws $48$ prompts from the listed sources, rolls out $24$ ODE trajectories per prompt, and fits Algorithm~\ref{alg:centered_regression} with the centers in Eq.~\eqref{eq:geneval_centers}. It trains the six lowest-reward trajectories of each prompt, restricted on GenEval and OCR prompts to trajectories with task reward below $0.99$. A bank update uses the PickScore configuration of Table~\ref{tab:config} with reward $10q$. The GenEval task reward is the partial-credit score up to update $450$ and the mean of the partial-credit and strict judgements afterwards. Updates $500$--$600$ use $40$ sampling steps with actor loss on every fourth state; all other stages use $10$ steps. The learning rate for updates $700$--$850$ is reached after a $20$-update warm-up. Table~\ref{tab:joint_stages} lists the stages.

\begin{table}[ht]
\caption{Stages of the five-reward joint model. The prompt mix gives the fractions of Pick-a-Pic, GenEval, and OCR prompts per update; the first stage is the first training seed of the single-reward PickScore run in Table~\ref{tab:single}.}
\label{tab:joint_stages}
\centering\small
\setlength{\tabcolsep}{4pt}
\begin{tabular}{llcl}
\toprule
Updates & Update type & Prompt mix & Learning rate\\
\midrule
0--200 & bank, PickScore & 1 / 0 / 0 & $3\times10^{-4}$\\
200--500 & group & 0.20 / 0.45 / 0.35 & $3\times10^{-4}$\\
500--600 & group & 0.20 / 0.60 / 0.20 & $1.35\times10^{-4}$\\
600--700 & bank & 1 / 0 / 0 & $7.25\times10^{-5}$\\
700--850 & group & 0 / 0 / 1 & $7.25\times10^{-5}$\\
850--1000 & group & 0.20 / 0.60 / 0.20 & $8.75\times10^{-5}$\\
\bottomrule
\end{tabular}
\end{table}

\paragraph{Compute.}
The full joint model uses \oMultiGPUh{} GPU-hours from SD3.5-Medium. DiffusionNFT does not report the cost of its joint model. Its described stages total $800+300+200+200+100=1{,}600$ iterations with $40$ sampling steps; its Table~1 lists $1.7$k, and we use the smaller count. Timing its official code on our hardware and accounting for the described schedule imply at least $2{,}000$ GPU-hours in total, or at least $5\times$ our budget.

\section{Measurement provenance}
\label{app:measurement}
Our single-reward and five-reward evaluations use the same protocols and datasets as DiffusionNFT, enabling direct comparison (Section~\ref{sec:setup}). Figure~\ref{fig:main} uses one image per DrawBench prompt for the PickScore curves, alongside the OCR and GenEval tests. Sampling counts can differ between checkpoint evaluations and learning-curve readouts, so their terminal values need not coincide.

Each curve retains its recorded cumulative GPU time. The source includes saved CSVs, a checksum manifest, and figure-generation scripts. Budget comparisons interpolate within observed runs; no missing result is extrapolated.

\section{Reward exploitation}
\label{app:hacking}
\subsection{Scheduled quality floor for OCR}
\label{app:quality_floor}
\begin{figure}[p]
\centering
\includegraphics[width=\linewidth]{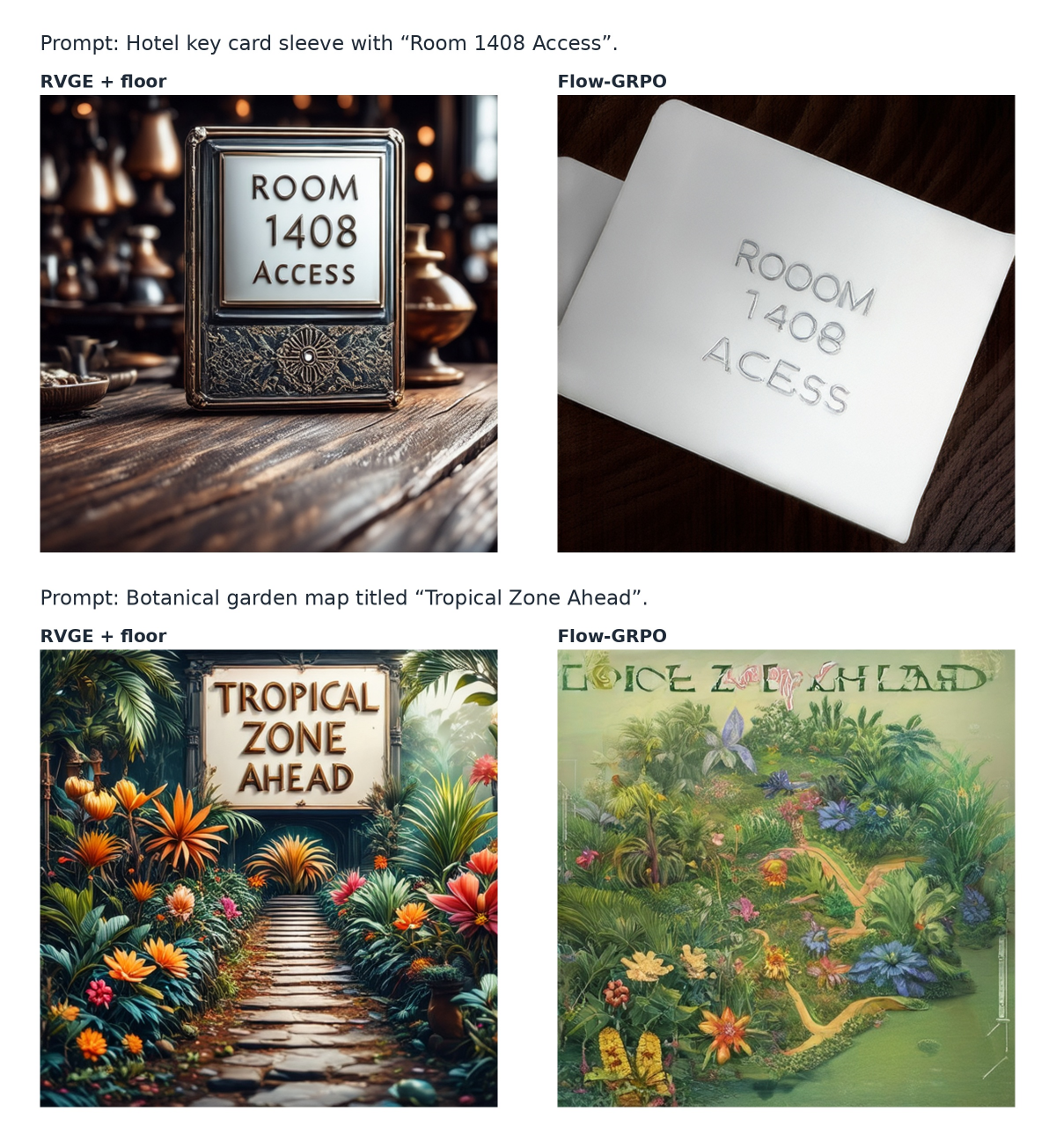}
\caption{OCR visual results comparing the \method quality-floor checkpoint with the public Flow-GRPO Text checkpoint on OCR test prompts. Within each pair, models use the same prompt, seed, and 40-step ODE sampler; panel labels abbreviate the prompts.}
\label{fig:qualitative_ocr}
\end{figure}

OCR-only training replaces requested scenes with large text on a flat background; on the main DrawBench evaluation, its PickScore is $\oOCRPick$ versus $22.34$ for the base model (Table~\ref{tab:single}). We counter this with a group-relative quality floor, using $q=\mathrm{PickScore}/26+\mathrm{ClipScore}+\mathrm{HPSv2.1}$ and regressing
\begin{equation}
 r=\mathrm{OCR}+\lambda_{\mathrm{floor}}\min(q-\bar q_c+m,0),\qquad \lambda_{\mathrm{floor}}=10,
 \label{eq:floor}
\end{equation}
where $\bar q_c$ is the prompt-group mean. A tight initial floor establishes scene content before relaxing the OCR tradeoff: $m=0$ until update $40$, $0.03$ until $100$, then $0.06$. After $150$ updates ($41$ GPU-hours), this run reaches OCR $\oOCRfOCR$ and DrawBench PickScore $\oOCRfPick$ while retaining scenes in the inspected samples (Figure~\ref{fig:qualitative_ocr}). The floor mitigates the observed flat-page exploit.

\paragraph{OCR layout comparison.}
We quantify templating by the mean cross-prompt correlation of $8\times8$ luminance thumbnails (base $0.05$; a fixed layout approaches one). At update $40$, the trajectory-bank and ODE-group runs in Section~\ref{sec:estimator_ablation} have correlations $0.93$ and $0.52$, respectively, and both produce flat, text-dominated layouts. The scheduled quality floor retains the requested scenes in the inspected samples, with a shared central-sign composition (correlation \oOCRfLayout). Its DrawBench Aesthetic score is $\oOCRfAes$, compared with $\oOCRAes$ for the OCR-only checkpoint (Table~\ref{tab:single}).

\paragraph{Why schedule the margin $m$?}
A high-noise reference penalty prevents the template but also removes the OCR signal. Training only failing trajectories still produces the flat-page layout, and a linear PickScore term preserves content but stalls OCR near $0.85$. A floor with a fixed reference ($\mathrm{PickScore}$ two points above the base model) preserves scenes and reaches $0.93$ after $120$ updates but stops there, because once the model's typical quality exceeds the reference the floor is silent; the relative floor of Eq.~\eqref{eq:floor} follows the model. Starting from initialization with the loose margin $m=0.06$ and $\lambda_{\mathrm{floor}}=10$ reaches OCR $0.95$ within $40$ updates, but on a cartoon-signboard template whose PickScore stays at the base level, so the quality rewards do not see it. The schedule starts with a tight margin to establish scene content, then relaxes it to improve OCR. The scheduled floor reaches \oOCRfOCR\ with scenes preserved; its remaining errors are mostly single glyphs in long strings, and the surviving style is a shared neon rendering of the text.

\begin{figure}[p]
\centering
\includegraphics[width=\linewidth]{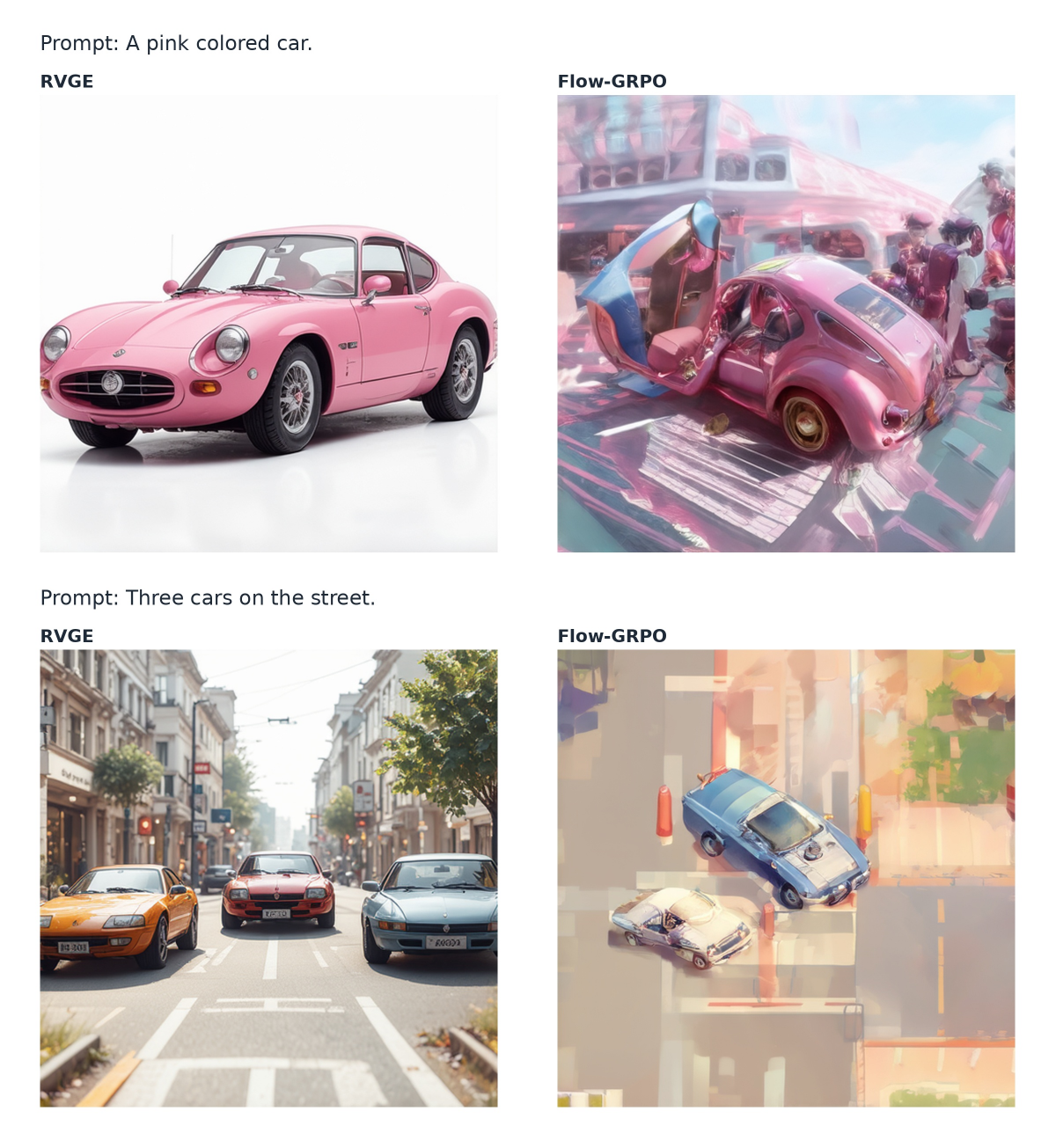}
\caption{PickScore visual results comparing the \method PickScore checkpoint with the public Flow-GRPO PickScore checkpoint on DrawBench prompts. Within each pair, models use the same prompt, seed, and 40-step ODE sampler.}
\label{fig:qualitative}
\end{figure}

\begin{figure}[p]
\centering
\includegraphics[width=\linewidth]{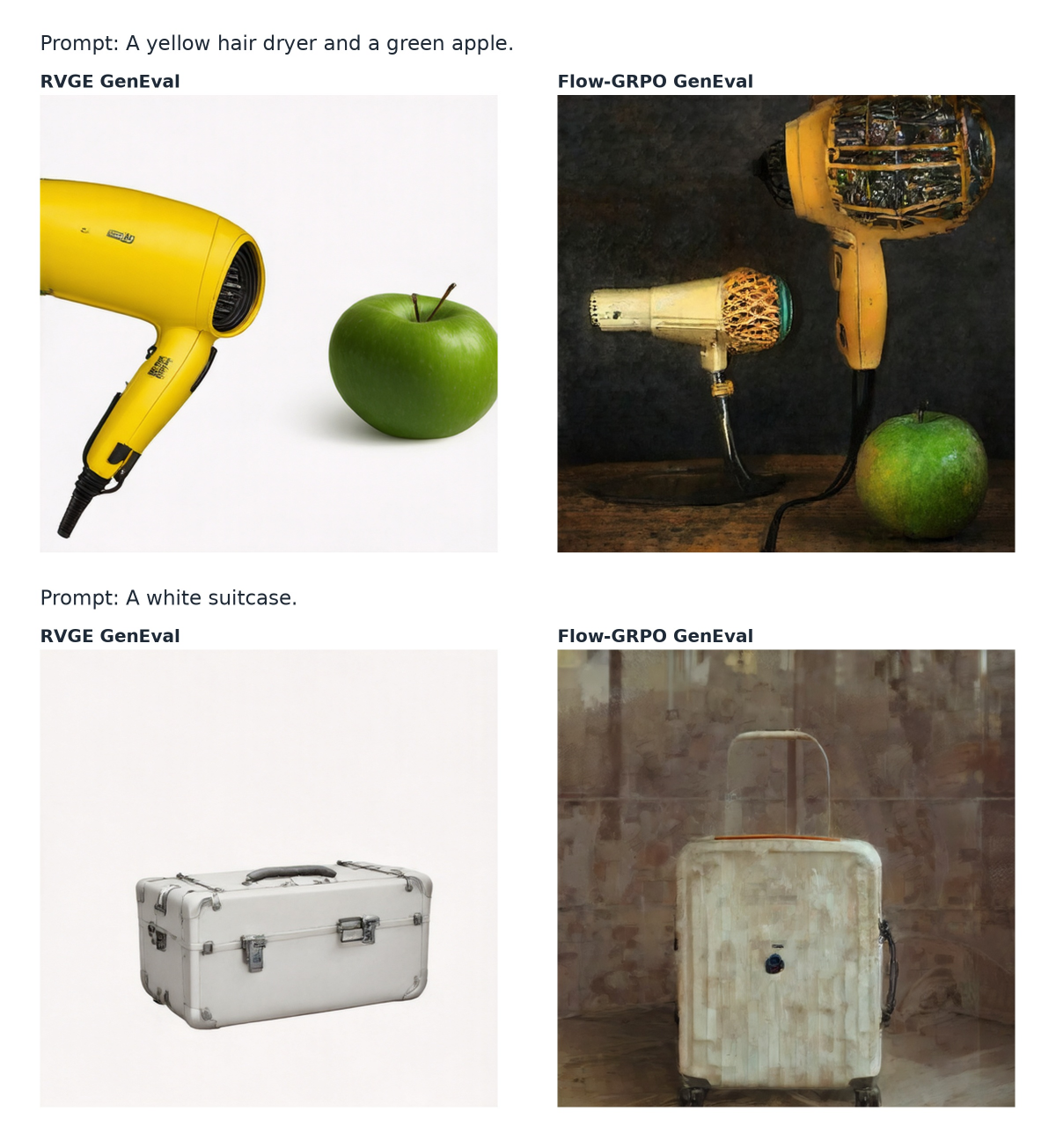}
\caption{GenEval visual results comparing the \method single-reward GenEval checkpoint with the public Flow-GRPO GenEval checkpoint on GenEval prompts. Within each pair, models use the same prompt, seed, and 40-step ODE sampler.}
\label{fig:qualitative_geneval}
\end{figure}

\begin{figure}[p]
\centering
\includegraphics[width=\linewidth]{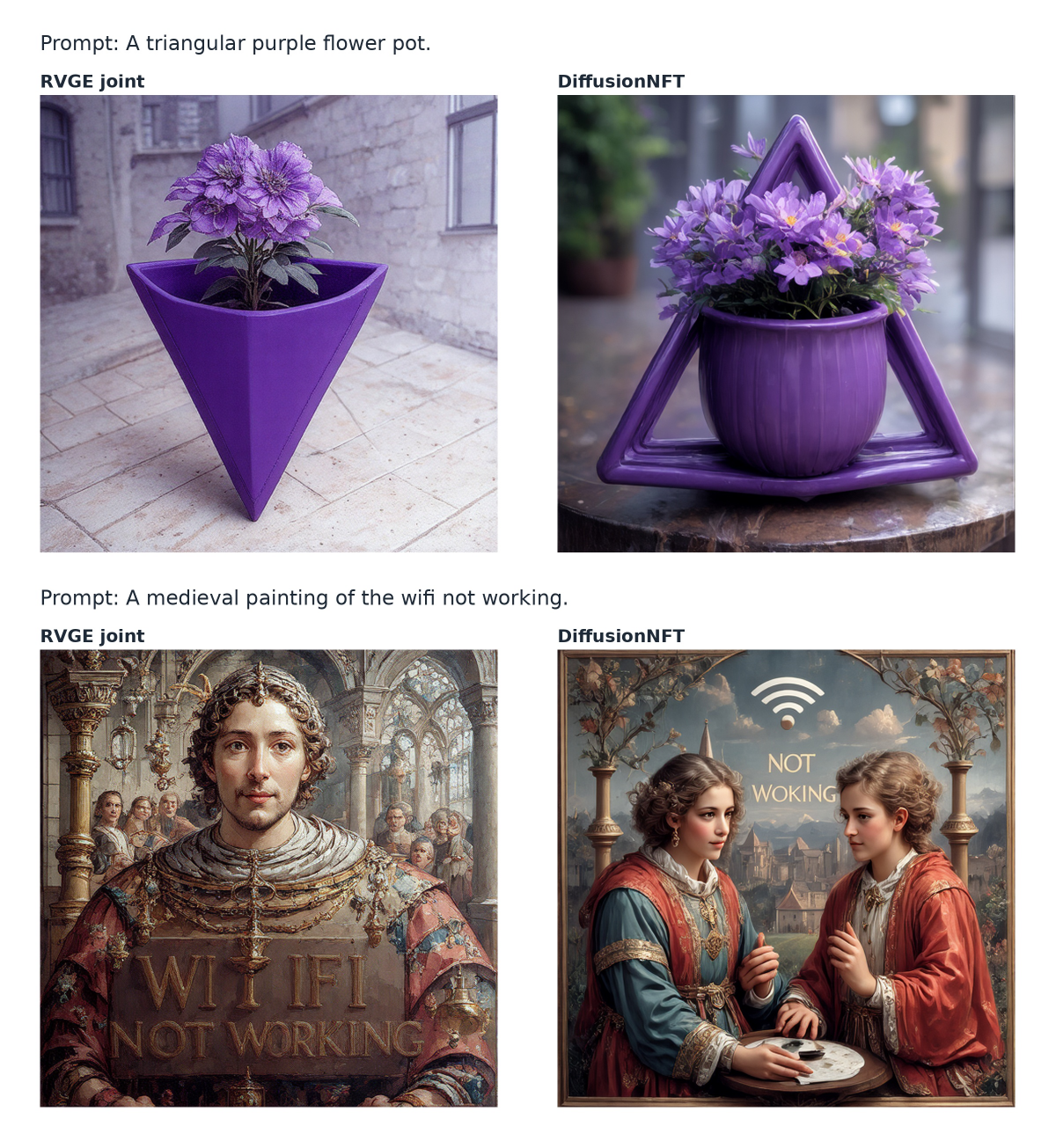}
\caption{DrawBench visual results comparing the five-reward joint \method checkpoint with the public DiffusionNFT multi-reward checkpoint. Each pair uses the same full prompt and seed with 40-step ODE sampling; panel labels abbreviate the prompts.}
\label{fig:qualitative_multireward}
\end{figure}

\begin{figure}[p]
\centering
\includegraphics[width=\linewidth]{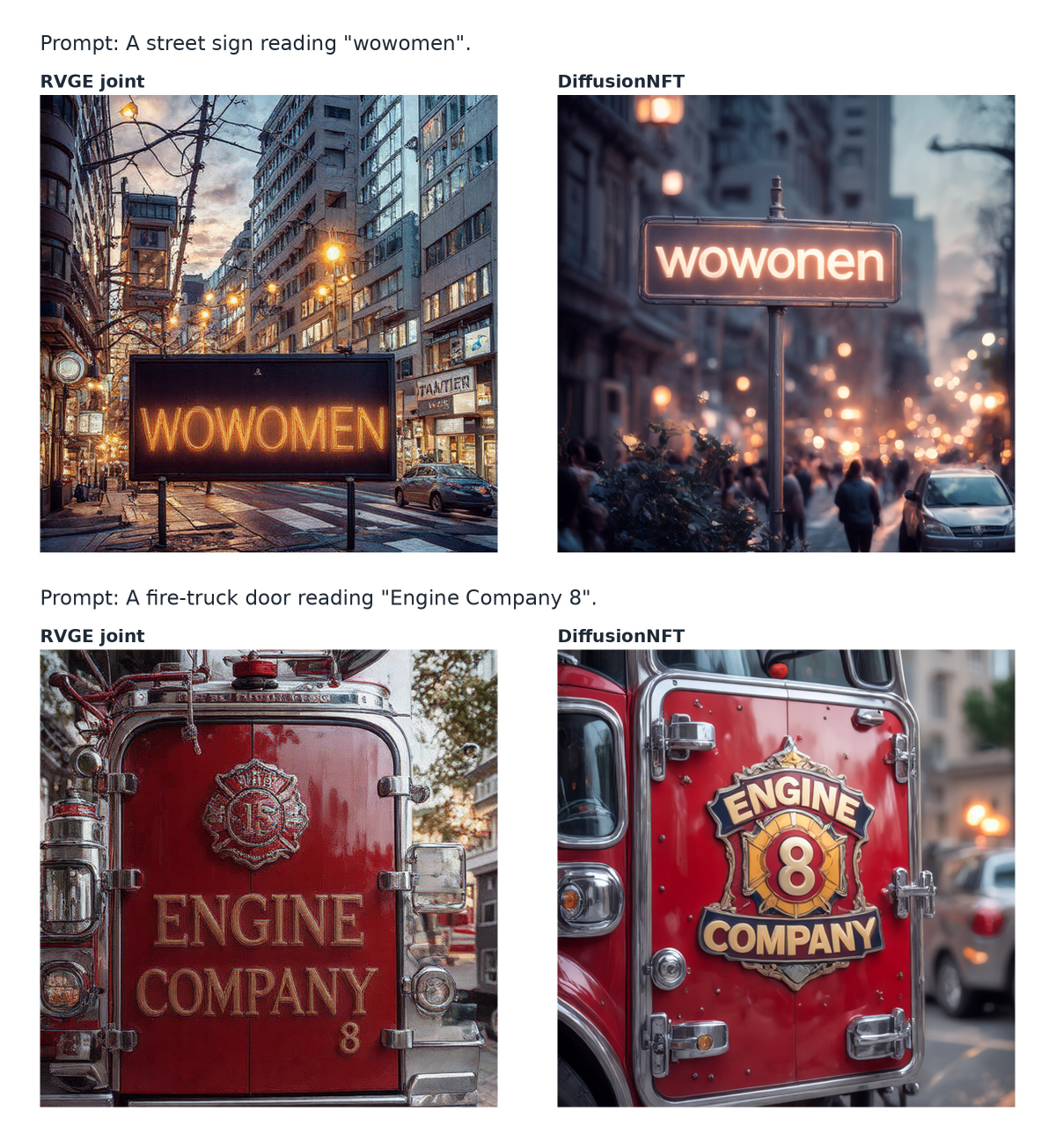}
\caption{OCR test visual results comparing the five-reward joint \method checkpoint with the public DiffusionNFT multi-reward checkpoint. Each pair uses the same full prompt and seed with 40-step ODE sampling; panel labels abbreviate the prompts.}
\label{fig:qualitative_multireward_ocr}
\end{figure}

\end{document}